\documentclass[10pt,twocolumn,letterpaper]{article}

\usepackage{template/cvpr}              

\usepackage[accsupp]{axessibility}  

\usepackage[dvipsnames]{xcolor}

\usepackage{bm}
\usepackage{tabularx}

\definecolor{cvprblue}{rgb}{0.21,0.49,0.74}
\usepackage[pagebackref,breaklinks,colorlinks,citecolor=cvprblue]{hyperref}

\def\paperID{13453} 
\def\confName{CVPR}
\def\confYear{2025}

\title{POp-GS: Next Best View in 3D-Gaussian Splatting with P-Optimality}

\author{Joey Wilson\\
University of Michigan\\
{\tt\small wilsoniv@umich.edu}
\and
Marcelino Almeida\\
Amazon Lab 126\\
{\tt\small mmalmeid@amazon.com}
\and
Sachit Mahajan\\
Amazon Lab 126\\
{\tt\small msachit@amazon.com}
\and
Martin Labrie\\
Amazon Lab 126\\
{\tt\small labrieml@amazon.com}
\and
Maani Ghaffari\\
University of Michigan\\
{\tt\small maanigj@umich.edu}
\and
Omid Ghasemalizadeh\\
Amazon Lab 126\\
{\tt\small ghasemal@amazon.com}
\and
Min Sun\\
Amazon Lab 126\\
{\tt\small aliensunmin@gmail.com}
\and
Cheng-Hao Kuo\\
Amazon Lab 126\\
{\tt\small chkuo@amazon.com}
\and
Arnab Sen\\
Amazon Lab 126\\
{\tt\small senarnie@amazon.com}
}

\begin{document}
\maketitle
\begin{abstract}
In this paper, we present a novel algorithm for quantifying uncertainty and information gained within 3D Gaussian Splatting (3D-GS) through P-Optimality. While 3D-GS has proven to be a useful world model with high-quality rasterizations, it does not natively quantify uncertainty or information, posing a challenge for real-world applications such as 3D-GS SLAM. We propose to quantify information gain in 3D-GS by reformulating the problem through the lens of optimal experimental design, which is a classical solution widely used in literature. By restructuring information quantification of 3D-GS through optimal experimental design, we arrive at multiple solutions, of which T-Optimality and D-Optimality perform the best quantitatively and qualitatively as measured on two popular datasets. Additionally, we propose a block diagonal covariance approximation which provides a measure of correlation at the expense of a greater computation cost. 

\end{abstract}

\begin{figure}
    \centering
    \begin{subfigure}{0.24\linewidth}
        \centering
        \includegraphics[trim={0 2cm 0 1cm},clip,width=\textwidth]{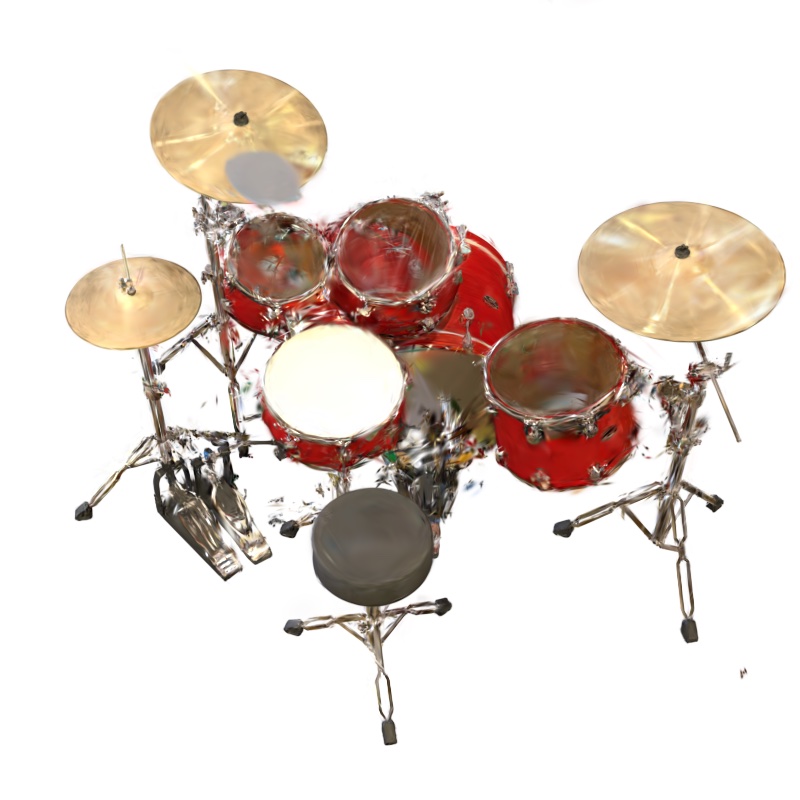}
    \end{subfigure}
    \begin{subfigure}{0.24\linewidth}
        \centering
        \includegraphics[trim={0 2cm 0 1cm},clip,width=\textwidth]{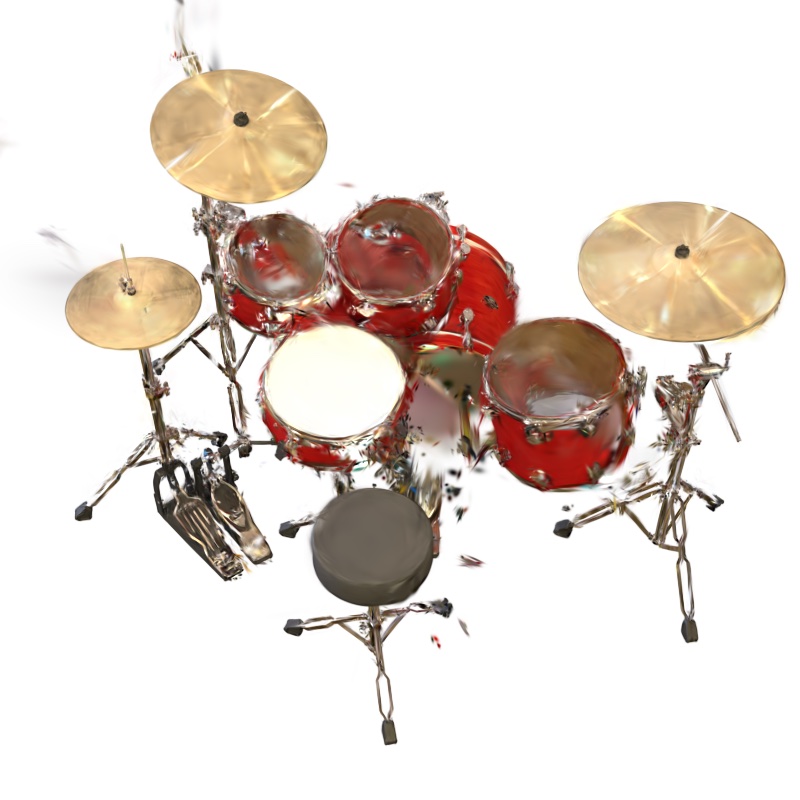}
    \end{subfigure}
    \begin{subfigure}{0.24\linewidth}
        \centering
        \includegraphics[trim={0 2cm 0 1cm},clip,width=\textwidth]{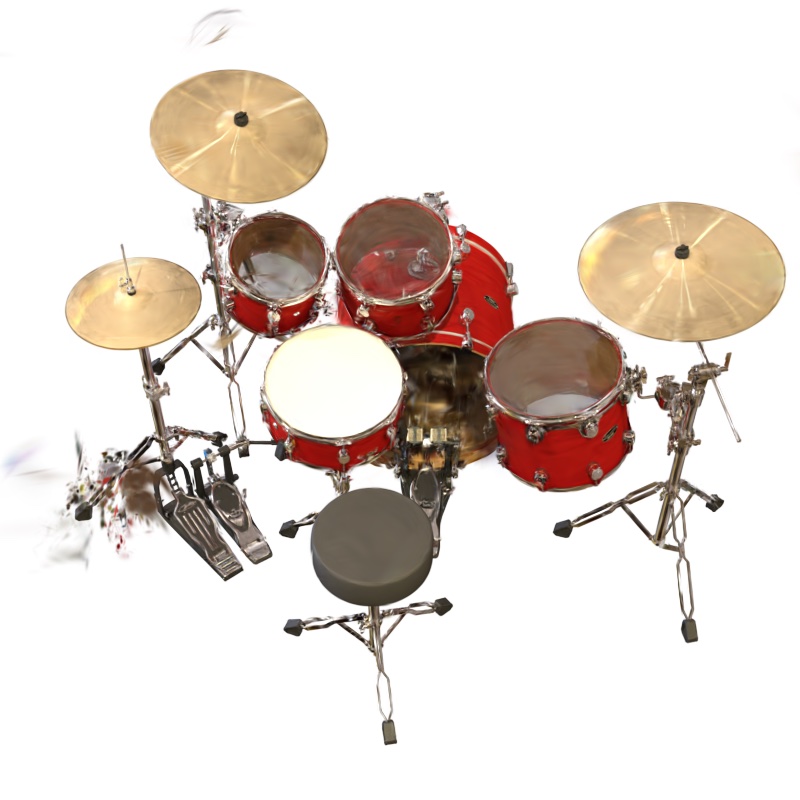}
    \end{subfigure}
    \begin{subfigure}{0.24\linewidth}
        \centering
        \includegraphics[trim={0 2cm 0 1cm},clip,width=\textwidth]{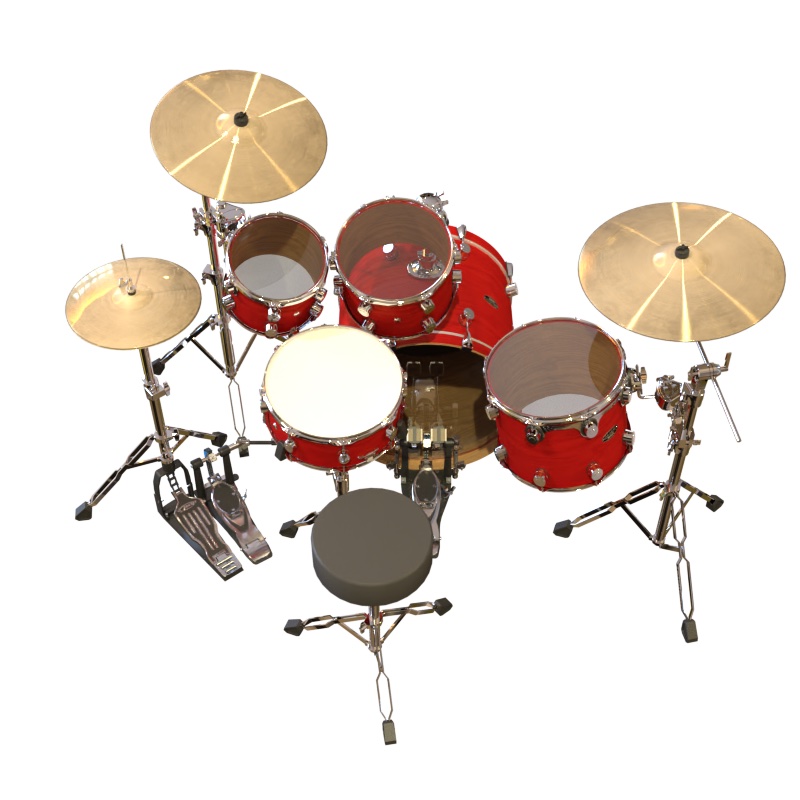}
    \end{subfigure}
    \begin{subfigure}{0.24\linewidth}
        \centering
        \includegraphics[trim={0 0 0 5cm},clip,width=\textwidth]{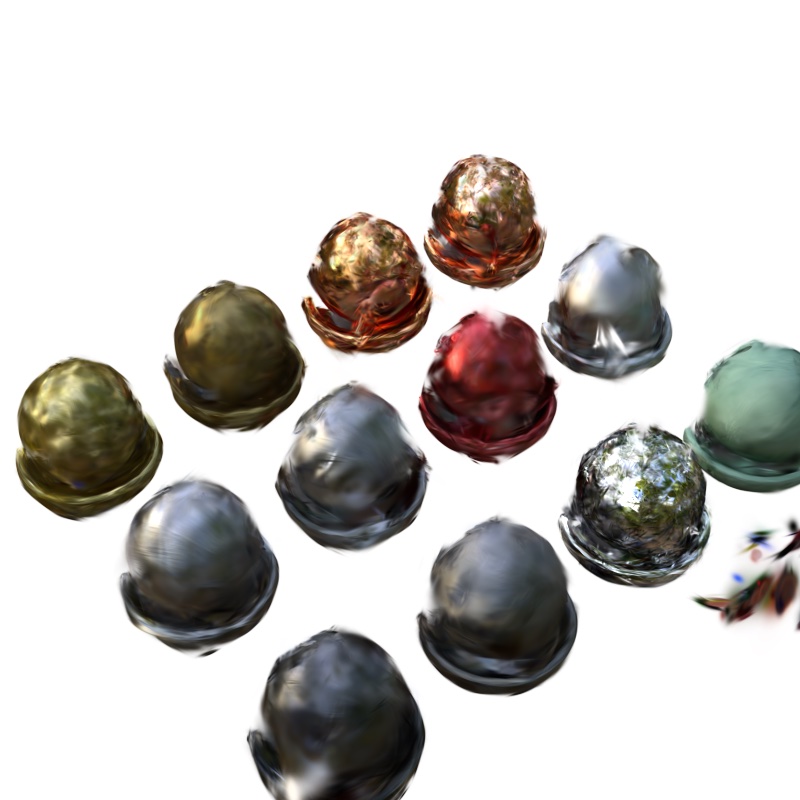}
    \end{subfigure}
    \begin{subfigure}{0.24\linewidth}
        \centering
        \includegraphics[trim={0 0 0 5cm},clip,width=\textwidth]{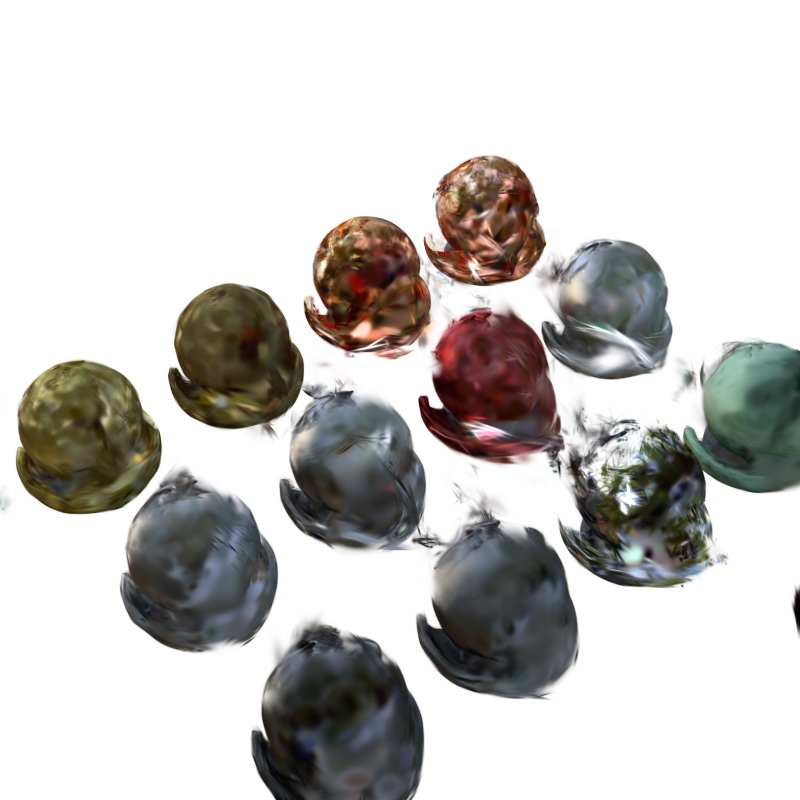}
    \end{subfigure}
    \begin{subfigure}{0.24\linewidth}
        \centering
        \includegraphics[trim={0 0 0 5cm},clip,width=\textwidth]{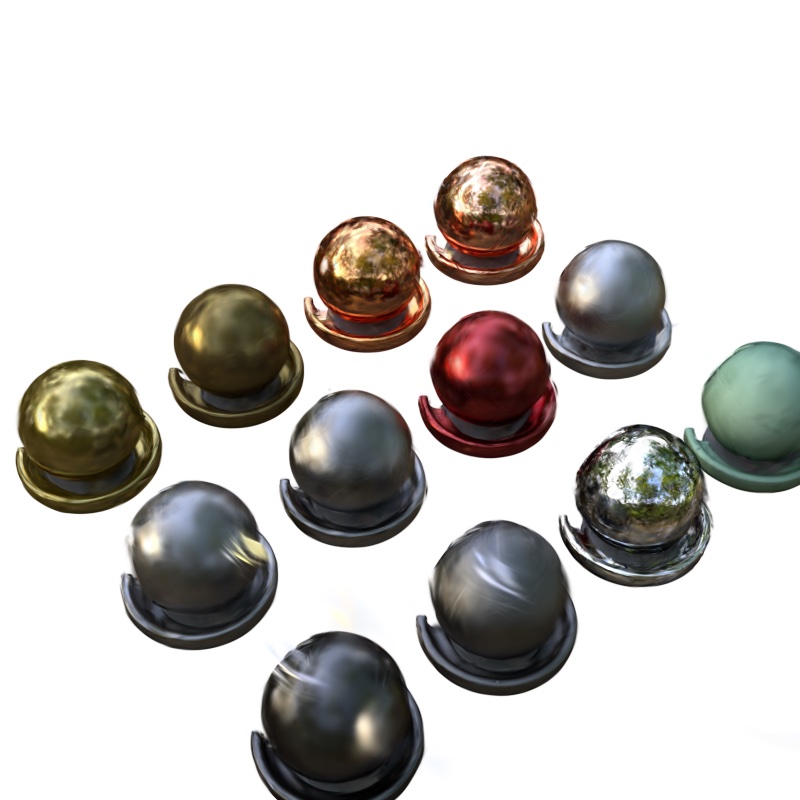}
    \end{subfigure}
    \begin{subfigure}{0.24\linewidth}
        \centering
        \includegraphics[trim={0 0 0 5cm},clip,width=\textwidth]{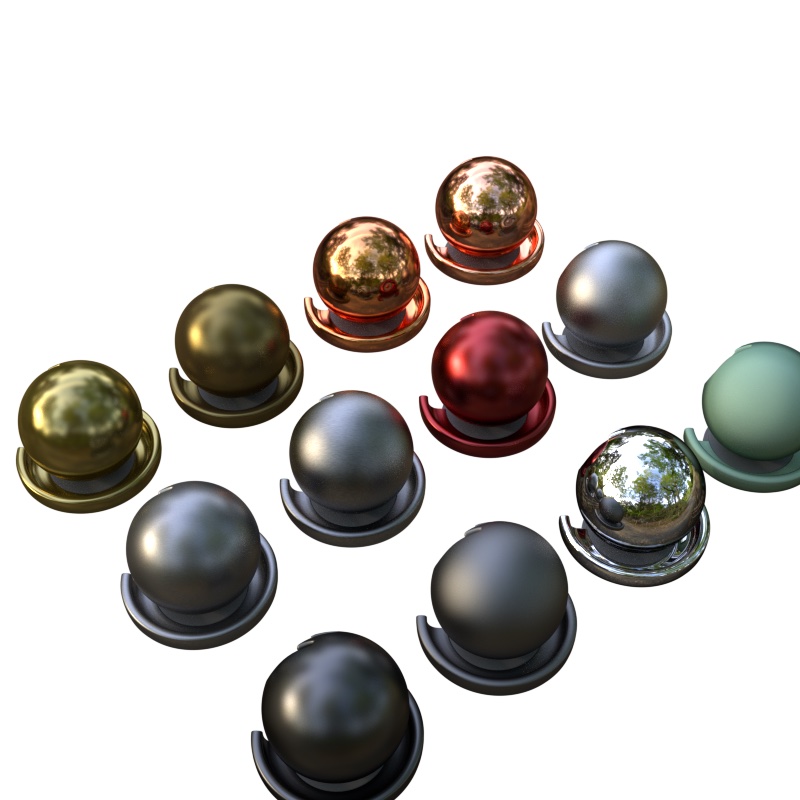}
    \end{subfigure}
    \begin{subfigure}{0.24\linewidth}
        \centering
        \includegraphics[trim={0 0 0 5cm},clip,width=\textwidth]{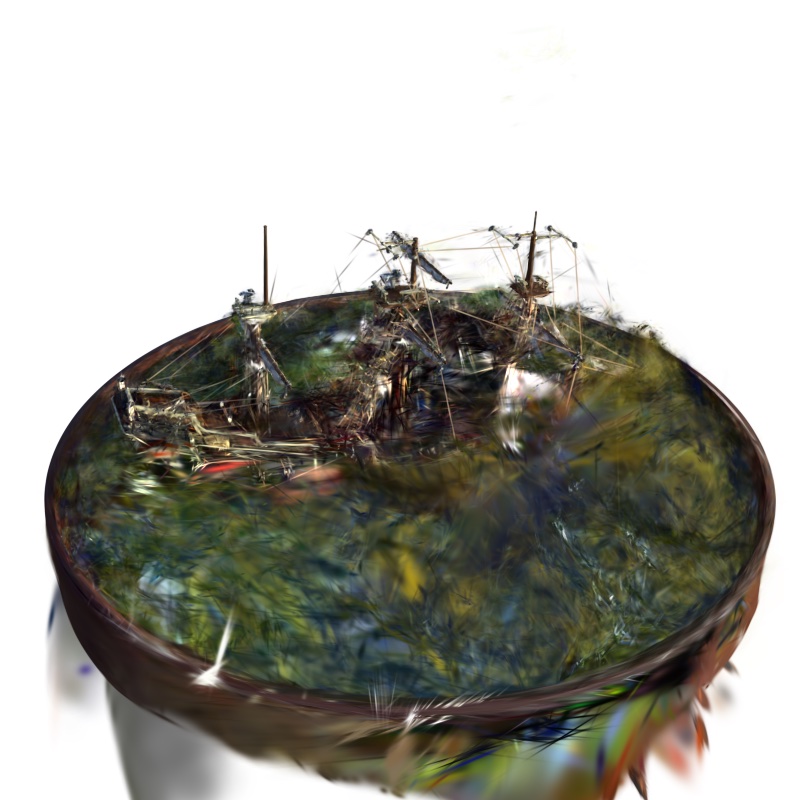}
        \caption{Uniform}
    \end{subfigure}
    \begin{subfigure}{0.24\linewidth}
        \centering
        \includegraphics[trim={0 0 0 5cm},clip,width=\textwidth]{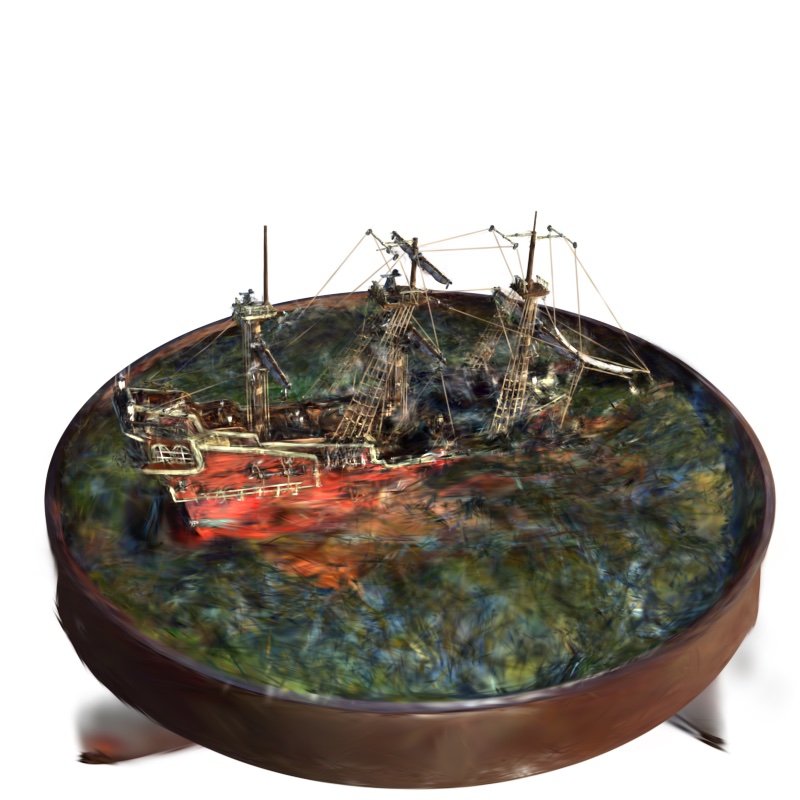}
        \caption{FisherRF}
    \end{subfigure}
    \begin{subfigure}{0.24\linewidth}
        \centering
        \includegraphics[trim={0 0 0 5cm},clip,width=\textwidth]{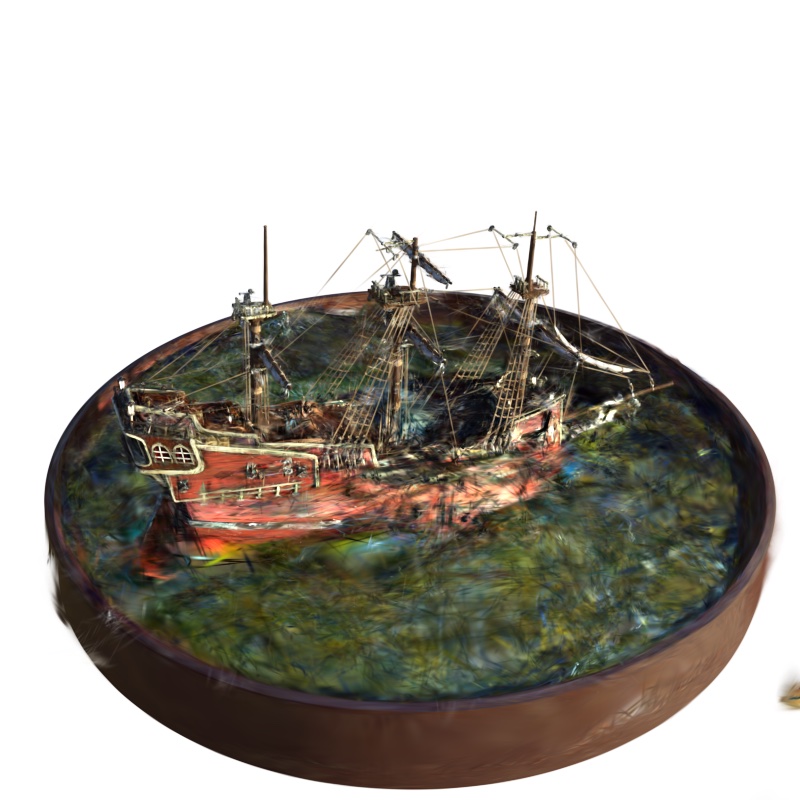}
        \caption{Ours}
    \end{subfigure}
    \begin{subfigure}{0.24\linewidth}
        \centering
        \includegraphics[trim={0 0 0 5cm},clip,width=\textwidth]{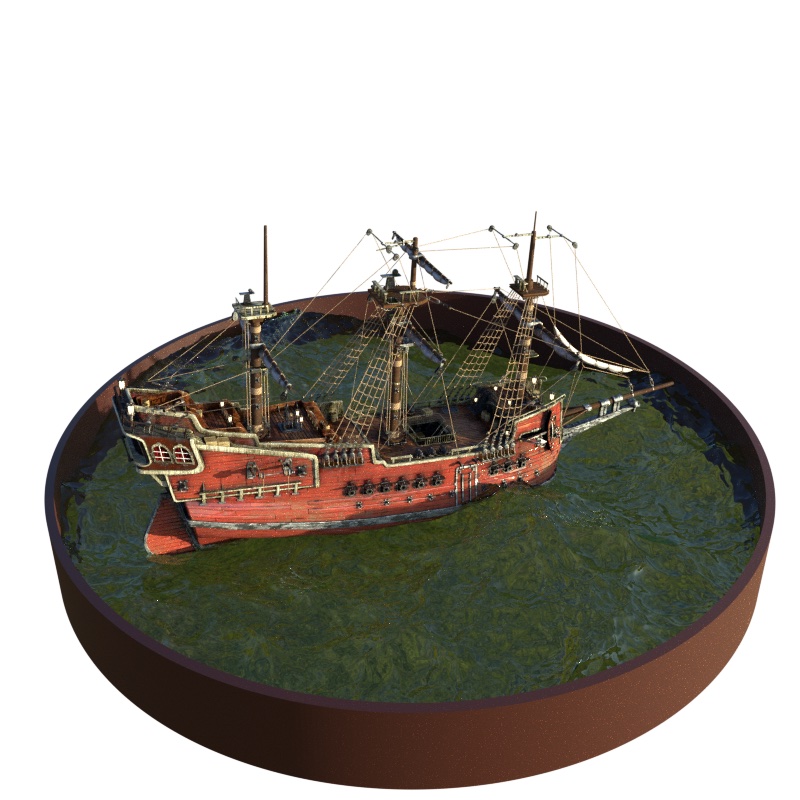}
        \caption{Ground Truth}
    \end{subfigure}
    \caption{We propose a novel method for calculating information gain from image in 3D Gaussian Splatting. In the above images, each method is provided a set of one hundred candidate views on scenes from the Blender dataset, and selects ten views to train a 3D-GS model on. Compared to state-of-the-art, our method more accurately estimate the information value of images to train a 3D-GS model. In this figure we demonstrate block D-Optimality, however our derivation also provides multiple solutions discussed later.}
    \vspace{-4mm}
    \label{fig:10_view_blender_comparison}    
\end{figure}

\section{Introduction}

In order to operate in a novel environment, robots must be capable of quantifying uncertainty in their surroundings due to occlusions and unobserved regions, as well as quantifying information gained by exploring new areas. While many prior methods for map representations were built on a metric space with natural uncertainty quantification, such representations result in discretization errors of the environment \cite{ConvBKIJournal}. Recently, 3D-Gaussian Splatting (3D-GS) was proposed as a method for high-quality novel-view synthesis \cite{3D-GS}, which captivated the attention of the robotics community due to its explicit world model representation, with the potential to function as a map \cite{3D-GS_Survey, 3D-GS_SLAM, 3D-GS_SLAM2, ContinuousSemanticSplatting}.

Given a set of views of a scene, 3D-GS learns to render novel views from any angle through gradient descent. 3D ellipsoids are iteratively fine-tuned, which can then be rasterized to a 2D image at novel views. Several works have shown that the 3D ellipsoids can be extended to include features beyond just color, such as the category of objects the ellipsoid represents, resulting in an improved level of semantic scene understanding \cite{Feature3D-GS, LangSplat}. However, 3D-GS does not natively quantify uncertainty, which leads to issues when determining whether an image has been seen before, or when quantifying the information gained by perceiving a new view. While several works have studied this problem for neural radiance fields (NeRF) \cite{BayesRays, ActiveNerf, NeRF}, the explicit representation of 3D-GS presents a unique challenge.

Since the conception of 3D-GS, several works have sought to quantify uncertainty directly upon a trained 3D-GS model \cite{UncertaintyPruning, FisherRF, Variational3DGS}. Of these works, one solution which has appeared effective is relating the per-pixel gradient of the 3D-GS parameters to the information or contribution of the parameters. Notably, FisherRF \cite{FisherRF} derived a solution for information quantification of views in 3D-GS through a diagonal approximation of Fisher Information \cite{FisherInformation}. FisherRF has since been applied to mobile manipulation \cite{FisherRFManipulation} and active perception \cite{FisherRFPerception}, however it does not leverage prior literature in effective functions for active perception \cite{kiefer1974general, sim2005global} or consider any correlation between parameters. 

Classically, the problem of quantifying information gain from views is known as active perception in robotics literature, and has been studied extensive in  probabilistic robotics applications \cite{ActiveSLAM, MaaniActiveMapping}. More generally, active perception can be solved through application of optimal experimental design \cite{ExperimentInformation}, which defines solutions for identifying the most informative design parameters for an experiment in order to learn about a set of unknown parameters \cite{MeasureInformation, Huan2013b, Huan2014a, KLD}. One particular solution to experimental design is the use of P-Optimality, which specifies a class of solutions depending on the covariance matrix and the choice of P. P-Optimality has been successfully applied in robotics to keyframe selection, optimal loop closure, and active perception, however has not been explored in 3D-GS \cite{placed2023surveyactivesimultaneouslocalization, placed2022explorb, carrillo2012onthecomparison}.    


Building upon the recent work of FisherRF, which shows that a diagonal approximation of the Hessian matrix can be effective when calculating information gain in 3D-GS, we derive a general solution for the covariance matrix in 3D-GS and apply optimal experimental design techniques. Since the covariance matrix is too large to store in memory, we propose a simple diagonal approximation following FisherRF \cite{FisherRF}, as well as a block diagonal approximation which captures cross-correlation of parameters within the same ellipsoid \cite{UncertaintyPruning}. From our general derivation, we construct and compare different p-optimality solutions, and find that D-Optimality and T-Optimality lead to significant improvements. We also find that our block diagonal approximation improves information gain quantification compared to the baselines. 

To summarize, our contributions are:

\begin{enumerate}
    \item Derive a general solution for the covariance matrix of 3D-GS, which allows application of optimal experimental design techniques.
    \item Propose several novel methods for quantifying the information value of images in 3D-GS. 
    \item Block-diagonal approximation of 3D-GS covariance matrix which captures correlation between parameters of the same ellipsoid.
    \item Quantitative and qualitative comparison of optimal experimental design solutions on 3D-GS to each-other, and to current state-of-the-art solutions.
\end{enumerate}

\section{Related Work}
In this section, we explore literature related to 3D Gaussian Splatting (3D-GS) and optimal experimental design for active perception. In this paper, we propose to leverage the theory of p-optimality from optimal experimental design literature to quantify information gain within 3D-GS.

\subsection{3D Gaussian Splatting}
3D Gaussian Splatting (3D-GS) is a new method for novel view synthesis which models scenes through 3D ellipsoids \cite{3D-GS}, different from previous methods which model scenes implicitly \cite{NeRF, BlockNerf}. Each scene is modeled through thousands or millions of 3D ellipsoids where each ellipsoid contain an opacity and color modeled by spherical harmonics to capture lighting effects. 3D ellipsoids are trained through gradient descent, and at inference time are rasterized to create 2D images at candidate views through a process known as ``splatting" \cite{3D-GS_Survey}.

Due to the explicit representation of scenes as 3D ellipsoids, 3D-GS has attracted a significant amount of attention from the computer vision and robotics research communities \cite{3D-GS_SLAM, 3D-GS_SLAM2, GS-SLAM}. 3D-GS has the potential to substitute as a more expressive world model representation, and many works have explored adding additional features such as from vision-language (VL) networks to create higher levels of scene understanding \cite{Feature3D-GS, LangSplat}. However, despite the name, 3D-GS does not provide a measure of uncertainty which limits applications in safety-critical or resource-constrained environments. Recently, several works have investigated uncertainty quantification and have found a promising research direction of relating Fisher Information to the explicit 3D-GS parameters \cite{FisherRF, UncertaintyPruning}. In particular, FisherRF developed a formulation for calculating information gain which treats the 3D-GS model as a black box, not requiring any additional training. However, FisherRF does not leverage any inter-parameter correlation when calculating the information of values to images, and does not utilize the rich literature of optimal experimental design and active perception which derive solutions for calculating information gain. 

\subsection{Active Perception}
Active perception is a well-studied problem in robotics literature which seeks to identify the optimal path to improve a map. Since capturing new views in the real world requires robot traversal, a significant amount of research has focused on optimal solutions to determining which view-points are most valuable. One successful approach is the application of P-optimality \cite{kiefer1974general}, which defines a class of optimal experimental design solutions based on functionals of the covariance matrix which vary with the choice of an integer $p$. P-Optimality (P-Opt.) has been widely and successfully applied to SLAM prior to the conception of 3D-GS in keyframe selection, optimal loop closure, and active perception \cite{placed2023surveyactivesimultaneouslocalization, placed2022explorb, carrillo2012onthecomparison}. 


Early research in optimal decision-taking for Simultaneous Localization and Mapping problems used T-optimality due to efficient computation as the trace of the covariance matrix \cite{sim2005global, mihaylova2003comparison}, eliminating the need to compute the eigenvalues. On the other hand, recent research has focused on D-optimality as a reliable metric for Optimal Experimental Design \cite{placed2023surveyactivesimultaneouslocalization, chen2020active, placed2022general, placed2022explorb}, especially in active mapping \cite{carrillo2012onthecomparison}. The recent success of D-optimality can be explained by its monotonicity property in active mapping scenarios \cite{rodriguez2018ontheimportance}, which guarantees that uncertainty increases monotonically as a robot moves through the scene. Additionally, from an information theoretic perspective, differential entropy of a multivariate Gaussian is proportional to the determinant of the covariance matrix, which is captured by D-optimality \cite{ElementsInformationTheory}. In this work, we propose to expand upon the formulation from FisherRF to incorporate parameter correlation and develop a more general solution which allows for application of classical optimal experimental design techniques.
\section{Method}
In this section, we introduce our method for quantifying uncertainty and information gain in 3D-Gaussian Splatting. First, we introduce preliminaries on the 3D Gaussian Splatting representation. Next, we describe our approximation of the covariance matrix for each ellipsoid, which provides a measure of uncertainty on the parameters. Finally, we detail our method for efficiently calculating the informational value of a candidate image. 

\subsection{Preliminaries: Gaussian Splatting} \label{sec:3D-GS_formulation}
3D Gaussian splatting represents a scene through volumetric rendering of optimized 3D ellipsoids. The geometry of each 3D ellipsoid is parameterized by a center $\mu$, scale $S$, and rotation $R$, while the color contribution of each ellipsoid is defined by opacity $\alpha$ and color $c$. Together, the rotation and scale define the shape of the 3D ellipsoid:

\begin{equation}
    \Sigma = R S S^T R^T.
\end{equation}

To render images, 3D ellipsoids are first splatted into 2D projections from a provided viewpoint, resulting in a 2D shape $\Sigma'$ and location $\mu'$. Next, the contribution $\alpha_n'$ of each 2D ellipsoid $n$ to pixel $x'$ is calculated through a kernel as:

\begin{equation}
    \alpha_n' = \alpha_n \times \text{exp}\left(  - \frac{1}{2} (x' - \mu_n')^T {\Sigma'}_n^{-1} (x' - \mu_n') \right).
\end{equation}

Finally, 2D ellipsoids are blended into pixels through a process known as alpha compositing, which computes the color of each pixel from a depthwise sorted list of Gaussians $\mathcal{N}$:

\begin{equation}
    C = \sum_{n=1}^{\mathcal{N}} c_n \alpha_n' \prod_{j=1}^{n-1} (1 - \alpha_j').
\end{equation}

Parameters are optimized by comparing the rendered image from a viewpoint with the ground truth image and performing gradient descent over a weighted loss function \cite{3D-GS, 3D-GS_Survey, SSIM}:

\begin{equation}
    \mathcal{L} = (1 - \lambda)\mathcal{L}_1 + \lambda \mathcal{L}_{\text{D-SSIM}},
\end{equation}

\noindent where $\lambda$ is a weighting function on the $\mathcal{L}1$ and structural similarity loss. Once the 3D-GS model is fitted to the training data, it can render views from any perspective, however it lacks any information on when the rendering may fail. In order to quantify the amount of information the fitted Gaussian Splatting model has on each parameter, we note that the $\mathcal{L}1$ loss is proportional to the partial derivative of the rendered pixel's color with respect to the parameters of the 3D-GS model, summed over all pixels in the training set. Based on this insight, we construct our covariance matrix to capture this information. 

\subsection{Information Gain through Optimal Experimental Design}
In order to quantify information gained through adding an image, we first require a measure of uncertainty, which is derived in this section. Approaching the problem from a maximum likelihood perspective, the maximum likelihood formulation aims to determine the solution variable $\bm{\theta}$ that minimizes the pixel error $\bm{e} = \bm{c} - \bm{h}(\bm{\theta})$ across all pixels in all frames, where $\bm{h}(\cdot)$ is the 3D-GS model described in Section~\ref{sec:3D-GS_formulation}. Assuming that all measured pixel errors are normal zero mean, independent and identically distributed (IID), i.e, $\bm{e} \sim \mathcal{N}(\bm{0}, \sigma_e \cdot \mathbb{I})$, then the maximum likelihood formulation seeks an optimal solution variable $\bm{\theta} \in \mathbb{R}^{l \times l}$ that maximizes the likelihood function:
\begin{gather}
    p(\bm{c} | \bm{\theta} ) = \exp \begin{pmatrix} - \frac{1}{2}\frac{\bm{e}^T \bm{e}}{\bm{\sigma}_e^2} \end{pmatrix}.
\end{gather}

Due to the monotonicity of the log function, maximizing the function above is equivalent to minimizing the log-likelihood function:

\begin{equation} \label{eq:neg_log_likelihood_function}
    - \sigma_e^2 \cdot \text{log} \, p(\bm{c} | \bm{\theta} ) = \frac{1}{2} \bm{e}^T \bm{e} 
\end{equation}

Assuming that we have an estimate of the solution variables $\bm{\theta}_*$, then we can expand the system's model in the vicinity of $\bm{\theta}_*$ using Taylor expansion as: $\bm{h}(\bm{\theta}) \approx \bm{h}(\bm{\theta}_*) + \bm{J} \Delta \bm{\theta}$, where $\Delta \bm{\theta} \triangleq \bm{\theta} - \bm{\theta}_*$ and $\bm{J} \triangleq \frac{\partial \bm{h}}{\partial \theta}|_{\bm{\theta} = \bm{\theta}_*}$. We can rewrite the residual function as:
\begin{align}
    \bm{e} &\approx \bm{c} - \bm{h}(\bm{\theta}_*) - \bm{J} \Delta \bm{\theta} \nonumber \\
             &= \bm{e}_* - \bm{J} \Delta \bm{\theta},
\end{align}
\noindent where the optimal residual is defined as $\bm{e}_{*} \triangleq \bm{c} - \bm{h}(\bm{\theta}_*)$. Substituting this in Eq.~\ref{eq:neg_log_likelihood_function}, we have that:
\begin{align} \label{eq:neg_log_likelihood_function_expansion}
    \frac{1}{2} \bm{e}^T \bm{e} \approx \frac{1}{2} \bm{e}_{*}^T \bm{e}_{*} - \bm{e}_{*}^T \bm{J} \Delta \bm{\theta} + \frac{1}{2} \Delta \bm{\theta}^T \bm{J}^T \bm{J} \Delta \bm{\theta}.
\end{align}
In order to satisfy the first order conditions for optimality in Eq.~\ref{eq:neg_log_likelihood_function_expansion}, it is necessary for its first order partial derivative w.r.t. the solution variables to be zero \cite{bar2004estimation}. This leads to the Gauss-Newton iterative optimization equation where $\Delta \bm{\theta}$ is updated as:
\begin{gather}
    \Delta \bm{\theta} = \begin{pmatrix} \bm{J}^T \bm{J} \end{pmatrix}^{-1} \bm{J}^T \bm{e}_*.
\end{gather}

\begin{figure}[t]
    \centering
    \includegraphics[width=0.8\linewidth]{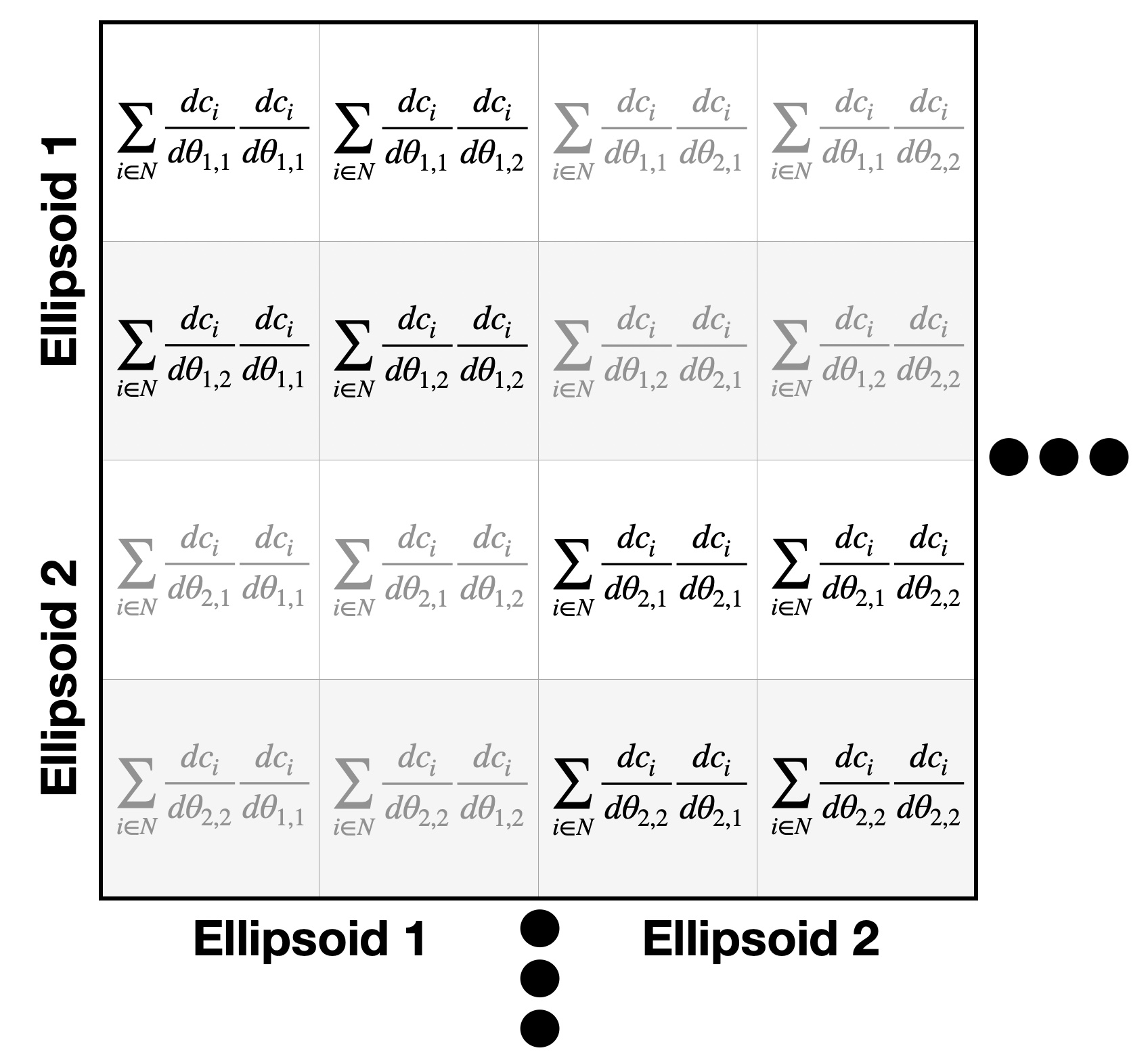}
    \caption{The Hessian matrix captures the information content of each parameter in the trained 3D-GS model, approximated as per-pixel gradients over the training set of images. Since a trained 3D-GS model may contain millions of parameters, we approximate the Hessian matrix through a block or main diagonal. }
    \vspace{-4mm}
    \label{fig:hessian_mat}
\end{figure}

Assuming that the optimal solution $\bm{\theta}_*$ is unbiased, then it follows that the expected residual $\mathbb{E}[\bm{e}_*] = \bm{0}$ and $\mathbb{E}[\bm{e}_*\bm{e}_*^T] = \bm{\sigma}_e \cdot \mathbb{I}$, leading to $\mathbb{E}[\Delta \bm{\theta}] = \bm{0}$ and:
\begin{align}
    \mathbb{E}[\Delta \bm{\theta} \Delta \bm{\theta}^T] 
        &= \begin{pmatrix} \bm{J}^T \bm{J} \end{pmatrix}^{-1} \bm{J}^T \mathbb{E}[\bm{e}_* \bm{e}_*^T] \bm{J} \begin{pmatrix} \bm{J}^T \bm{J} \end{pmatrix}^{-1} \nonumber \\
        &= \sigma_e^2 \begin{pmatrix} \bm{J}^T \bm{J} \end{pmatrix}^{-1} \bm{J}^T \bm{J}^T \bm{J} \begin{pmatrix} \bm{J}^T \bm{J} \end{pmatrix}^{-1} \nonumber \\
        &= \sigma_e^2 \begin{pmatrix} \bm{J}^T \bm{J} \end{pmatrix}^{-1}.
\end{align}
Without loss of generality, this work assumes\footnote{Given that we've assumed that the pixel error is zero-mean IID, the value of $\sigma_e$ does not matter for this application.} $\sigma_e = 1$. Defining $\bm{H} \triangleq \bm{J}^T \bm{J} \in \mathbb{R}^{l \times l}$, the matrix $\bm{H}$ is known as an approximation of the Hessian, or the information matrix for this nonlinear optimization problem. Therefore, the covariance matrix associated with the solution variables $\bm{\theta}$ is given by the inverse of the Hessian (information) matrix.

\begin{table*}[ht]
\centering
\caption{Properties of P-Optimality for different values of $p$. Note: $\lambda_k$ represent the eigenvalues of the covariance matrix $\bm{\Sigma}_i$.}
\begin{tabularx}{\textwidth}{|>{\centering\arraybackslash}p{1.5cm}|>{\centering\arraybackslash}p{3.2cm}|>{\centering\arraybackslash}p{3.5cm}|>{\centering\arraybackslash}X|>{\centering\arraybackslash}p{3.3cm}|} \hline
            & \textbf{T-optimality} & \textbf{A-optimality} & \textbf{D-optimality} & \textbf{E-optimality} \\ \hline
\textbf{p}  & 1            & -1           & 0            & $\mp \infty$ \\ \hline
\textbf{Equivalent Formulae}
            & $\frac{1}{l} \text{tr}\begin{pmatrix} \bm{\Sigma}_i \end{pmatrix} = \frac{1}{l} \sum_k^{l} \lambda_k$
            & $\left( \frac{1}{l} \text{tr}\begin{pmatrix} \bm{H}_i \end{pmatrix} \right)^{-1} = \left( \frac{1}{l} \sum_k^{l} \lambda_k^{-1} \right)^{-1}$
            & $\sqrt[l]{|\bm{\Sigma}_i|} = \exp \begin{pmatrix} \frac{1}{l} \sum_k^{l} \log \lambda_k \end{pmatrix}$
            & \begin{tabular}[c]{@{}l@{}} $\min_{\lambda_k}$ \\ $\max_{\lambda_k}$ \end{tabular} \\ \hline
\textbf{Meaning}     & Average Variance & Harmonic Mean Variance & Volume of covariance hyper-ellipsoid & Single extreme eigenvalue \\ \hline
\end{tabularx}
\label{table:p-optimality}
\vspace{-4mm}
\end{table*}

\subsubsection{Uncertainty Decrease due to an Added Image}\label{sec:uncertainty_relationship}

In this paper, we assume that we have a set of $n$ images to choose from (along with their respective original poses) to determine which of the images will lead to maximal uncertainty reduction among all $n$ candidates. Therefore, our Next-Best-View formulation attempts at maximally decreasing uncertainty of the covariance matrix $\bm{\Sigma}_i$ that is obtained by adding the i-th image to the model, $i \in \{1, \cdots, n\}$. Note that the contents of the i-th image are not necessary for our formulation, only the pose at which we plan to take the i-th image from. This is an important feature of our solution, as it attempts to evaluate the amount of uncertainty reduction (or information increase) that can be achieved by taking an image from a new perspective without actually having the image available.

In this section, we assume that we already have an initial guess of the map $\bm{\theta}_*$ and that its associated prior Jacobian $\bm{J}_{-}$ and Hessian $\bm{H}_{-} = \bm{J}^T_{-} \bm{J}_{-}$ have been computed. As we add one new prospective candidate image $i$ taken from a pose $\bm{p}_i$, it is possible to compute the Jacobian associated with the new image using prior map parameters $\bm{\theta}_*$ and the image's pose $\bm{p}_i$ as $\bm{J}_i = \frac{\partial \bm{h}}{\partial \bm{\theta}}|_{\bm{\theta}_*, \bm{p}_i}$. Defining $\bm{H}_i$ as the Hessian of the problem as we add the i-th image, it can be computed as:
\begin{gather}
    \bm{H}_i = \bm{H}_{-} + \bm{J}_i^T \bm{J}_i.
\end{gather}

\begin{figure}[t]
    \centering
    \includegraphics[width=0.8\linewidth]{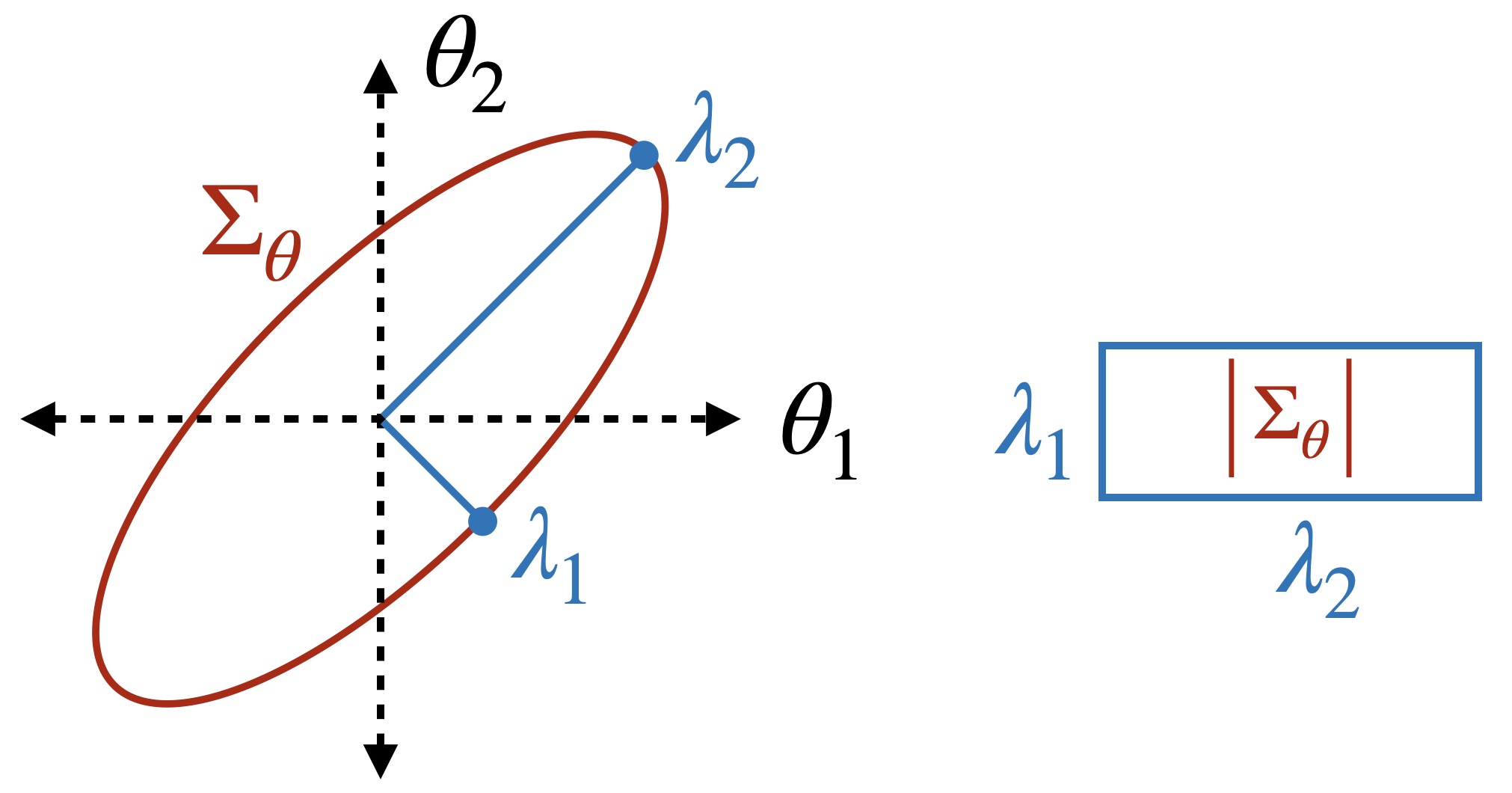}
    \caption{Optimal experimental design defines functionals of the eigenvalues of the covariance matrices, each with geometric intuitions. D-Optimality approximates the volume of the covariance matrix, as shown in this figure.}
    \label{fig:d_optimality}
    \vspace{-4mm}
\end{figure}

For exponential likelihoods, there is an asymptotic inverse relationship in the maximum likelihood estimator between the Hessian and covariance matrices \cite{AsymptoticStatistics}. Therefore, our goal is to maximize the information  $\mathcal{I}\begin{pmatrix} \cdot \end{pmatrix}$ obtained from the i-th image, which can be found by minimizing the uncertainty  $U\begin{pmatrix} \cdot \end{pmatrix}$:


\begin{align}
    \text{arg}\max_{i} \mathcal{I} \begin{pmatrix} \bm{H}_i \end{pmatrix}  &= \text{arg}\min_{i} U \begin{pmatrix} \bm{\Sigma}_i \end{pmatrix} \nonumber \\
    &= \text{arg}\min_{i} U \begin{pmatrix} \bm{H}_i^{-1} \end{pmatrix}.
\end{align}

In this work, we rely on the Theory of Optimal Experimental Design \cite{placed2023surveyactivesimultaneouslocalization, kiefer1974general}, which defines the P-Optimality uncertainty metric as:
\begin{gather}
    U_p(\bm{\Sigma}_i) = \begin{pmatrix} \frac{1}{l} \text{trace} \begin{pmatrix} \bm{\Sigma}_i^p \end{pmatrix} \end{pmatrix}^{\frac{1}{p}},
\end{gather}
\noindent where $p$ is an integer. Depending on the chosen value for $p$, the uncertainty function can have some special properties \cite{placed2022general}, as detailed in Table~\ref{table:p-optimality}.

\subsection{Approximating the Covariance}
In practice fitted 3D-GS models may contain millions of parameters, which is intractable due to the cubic computational and quadratic memory complexity of computing the eigenvalues. Therefore, we propose two approximations of the Hessian matrix to save memory and computation.

\textbf{Simple Diagonal:} First, following the work of FisherRF we propose a simple diagonal approximation of the covariance matrix. FisherRF derives the approximation from a Laplace approximation \cite{LaplaceApprox}, where the covariance matrix is approximated as the main diagonal plus a small regularizing constant $\lambda_\theta$: 

\begin{equation}
    \Sigma_\theta \approx \text{diag}(\Sigma_i) + \lambda_\theta,
\end{equation}

\noindent This formulation allows for an efficient and direct comparison of our method of computing information gain with FisherRF, without consideration of correlation. Intuitively, the constant $\lambda_\theta$ represents prior information for our method. 

\textbf{Block Diagonal:} In order to capture some of the correlation between 3D-GS parameters, we propose to approximate the Hessian matrix as a block diagonal matrix where each block diagonal element contains the parameters of a single ellipsoid. Please note that a block diagonal approximation has also been proposed within 3D-GS, however the approximation was applied to the task of pruning 3D-GS models to remove redundant ellipsoids \cite{UncertaintyPruning}. Our insight is that the parameters of the same ellipsoid are most likely to be correlated, and a block diagonal matrix can be processed in parallel on a GPU for efficient computation. 

When constructing the block diagonal matrix, we note that computing partial derivatives w.r.t. each pixel leads to singularity issues since the partial derivative of each color is the value $\alpha'_n$ for the ellipsoid. To avoid singularity issues, we therefore compute the Hessian matrix by separately calculating partial derivatives for each channel of every pixel, instead of for every pixel for all channels. The block diagonal approximation is shown in Fig. \ref{fig:hessian_mat}.

\subsection{Batch Selection}
In practical applications it may be valuable to measure information gain over a set of candidate images, such as along a trajectory or in keyframe selection. Following FisherRF, we implement a simple approach which iteratively adds or removes the most optimal candidate image, updates the Hessian, and repeats the process without additional training. While simple, this batch information implementation does capture the redundancy of views, as the change in the Hessian is reflected when a candidate image is added.

\begin{figure*}[t]
    \centering
    \begin{subfigure}{0.22\textwidth}
        \centering
        \includegraphics[width=\textwidth]{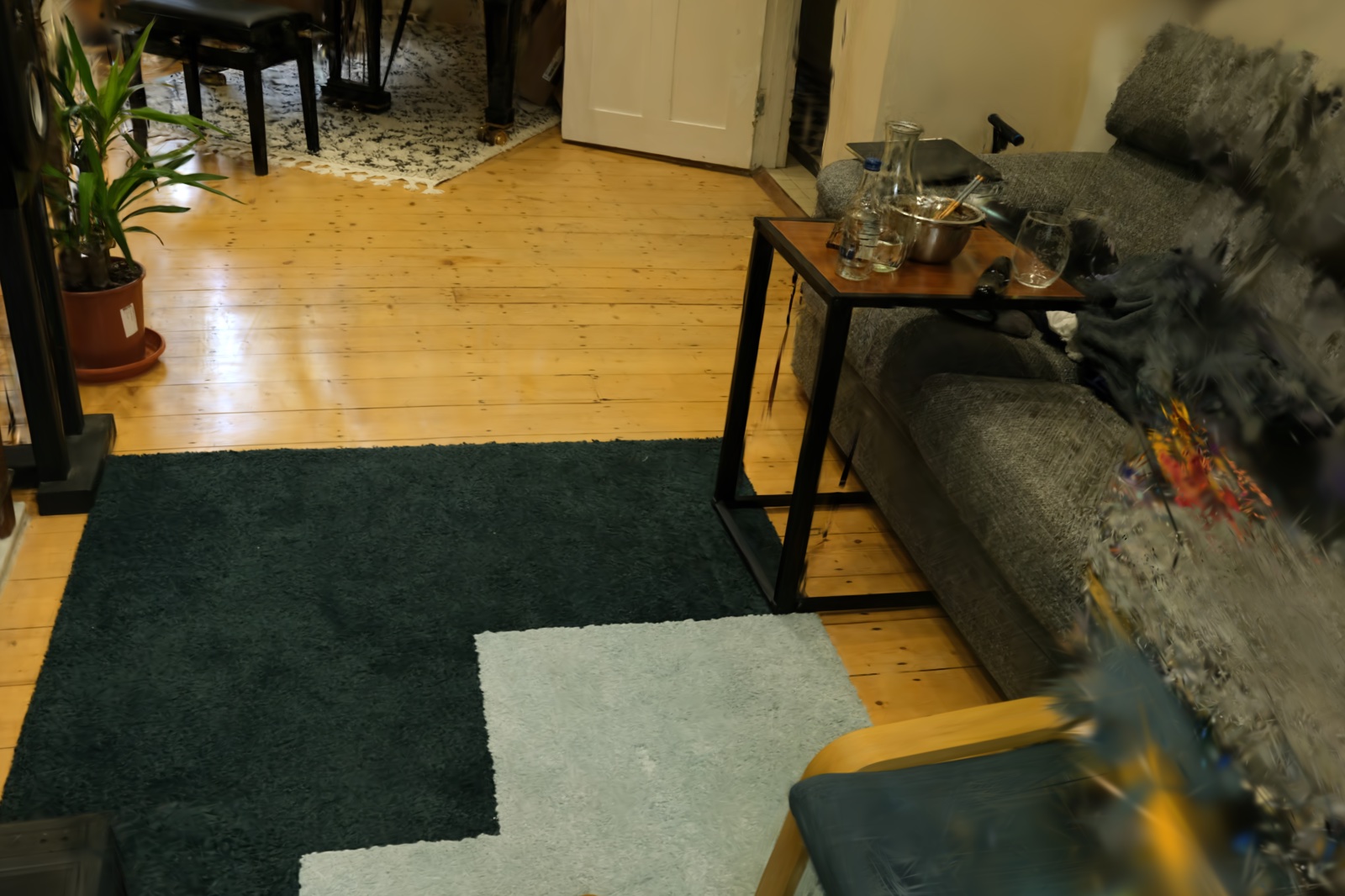}
    \end{subfigure}
    \begin{subfigure}{0.22\textwidth}
        \centering
        \includegraphics[width=\textwidth]{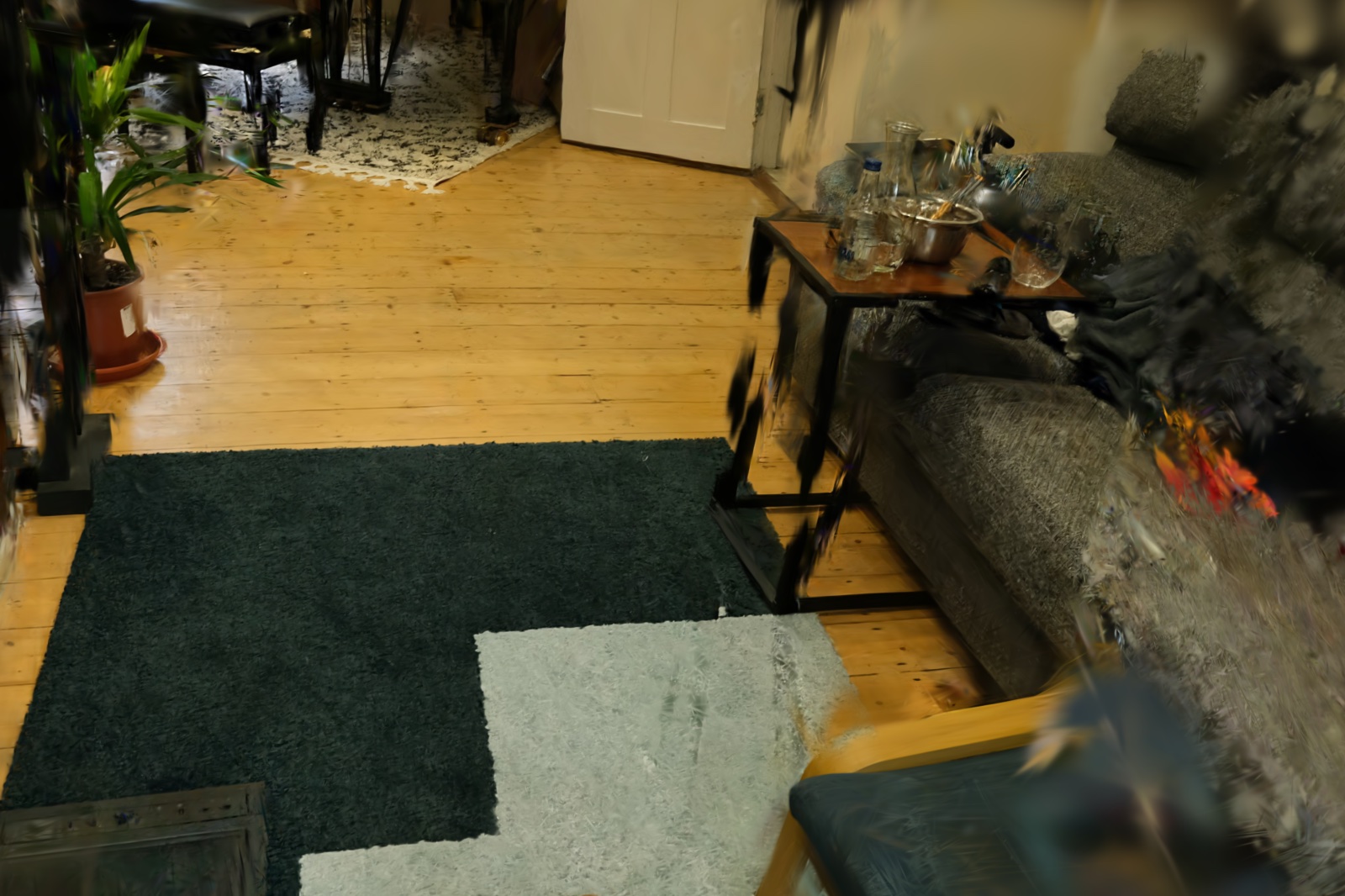}
    \end{subfigure}
    \begin{subfigure}{0.22\textwidth}
        \centering
        \includegraphics[width=\textwidth]{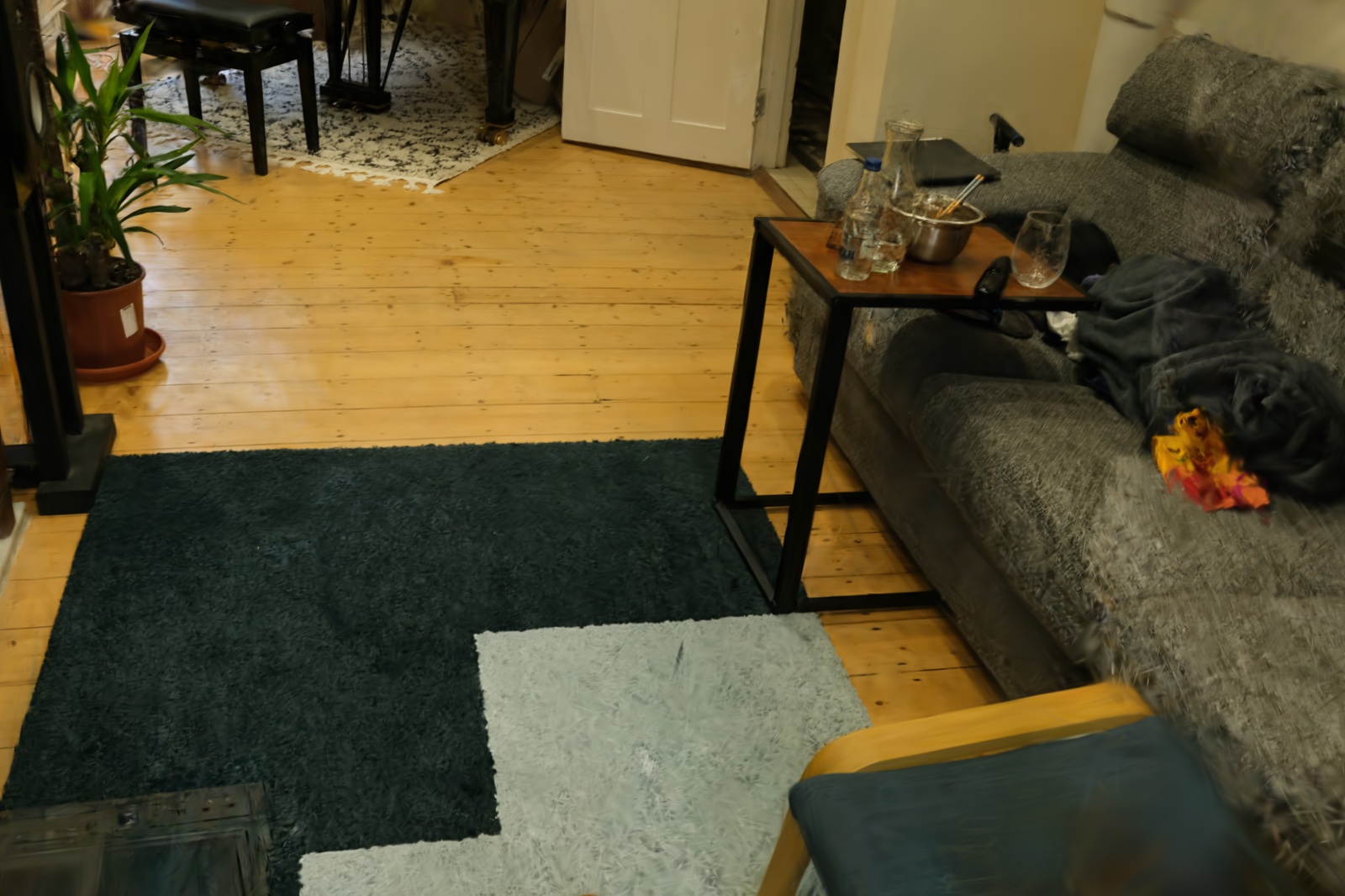}
    \end{subfigure}
    \begin{subfigure}{0.22\textwidth}
        \centering
        \includegraphics[width=\textwidth]{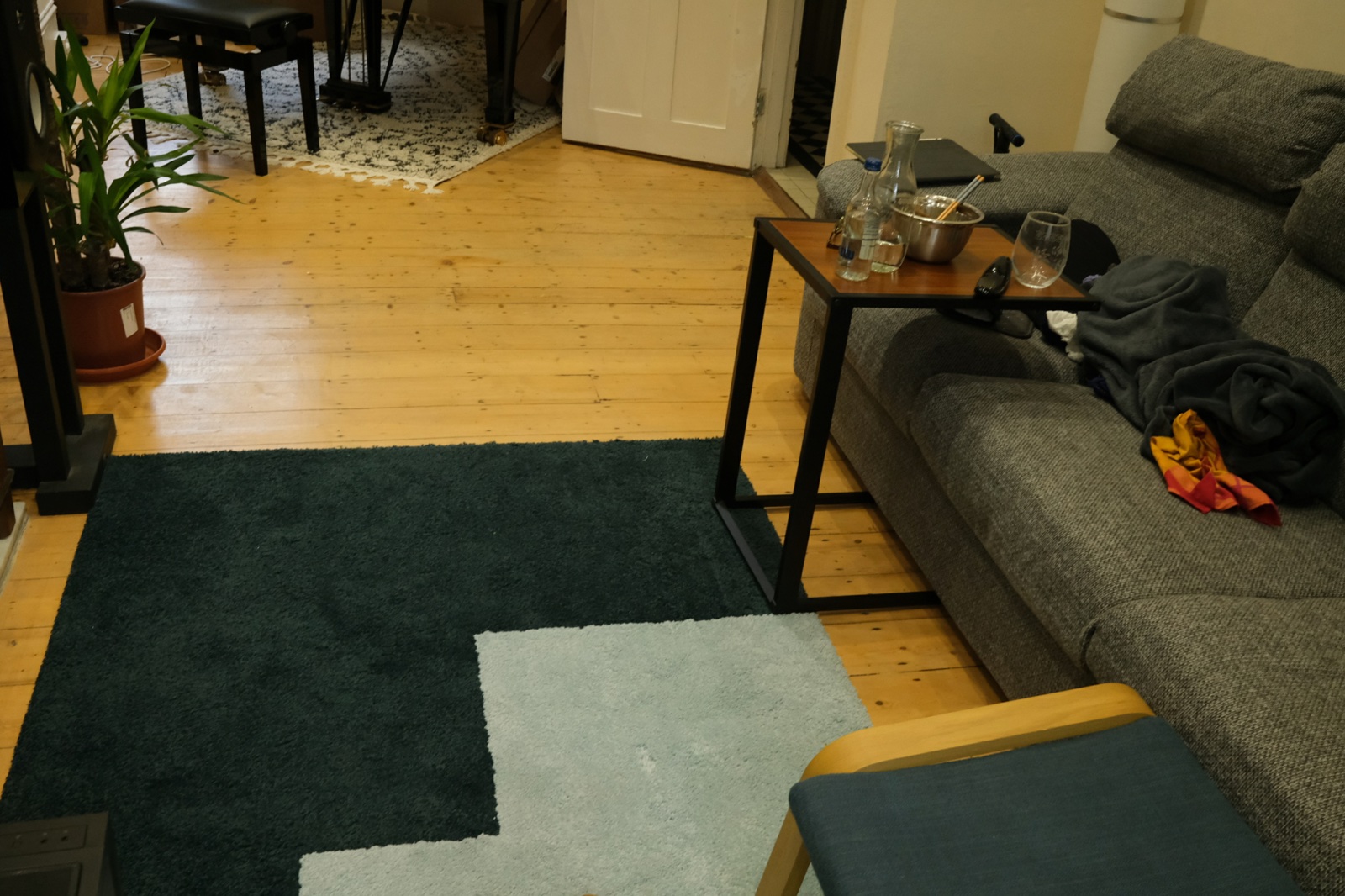}
    \end{subfigure}
    \begin{subfigure}{0.22\textwidth}
        \centering
        \includegraphics[width=\textwidth]{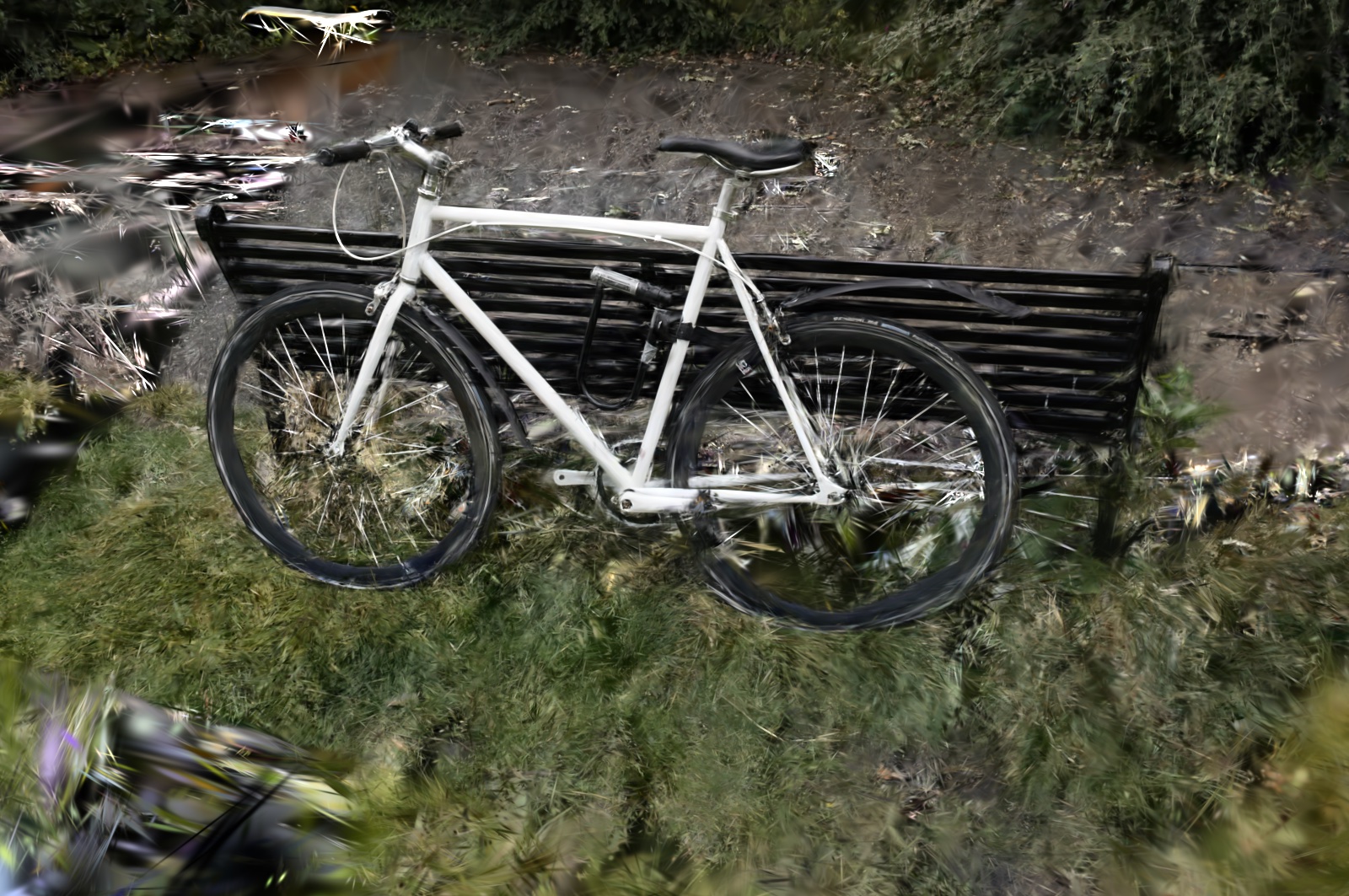}
    \end{subfigure}
    \begin{subfigure}{0.22\textwidth}
        \centering
        \includegraphics[width=\textwidth]{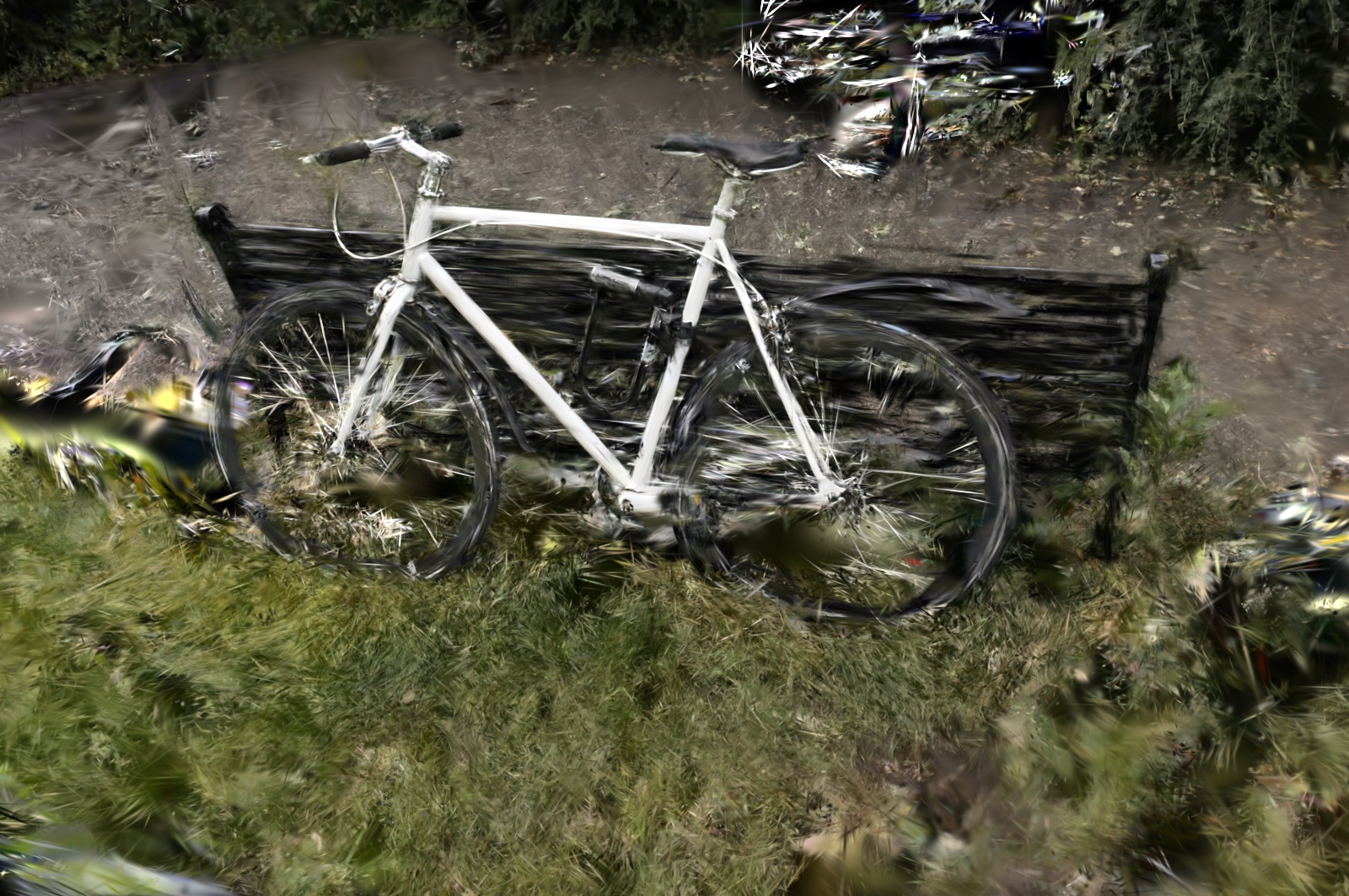}
    \end{subfigure}
    \begin{subfigure}{0.22\textwidth}
        \centering
        \includegraphics[width=\textwidth]{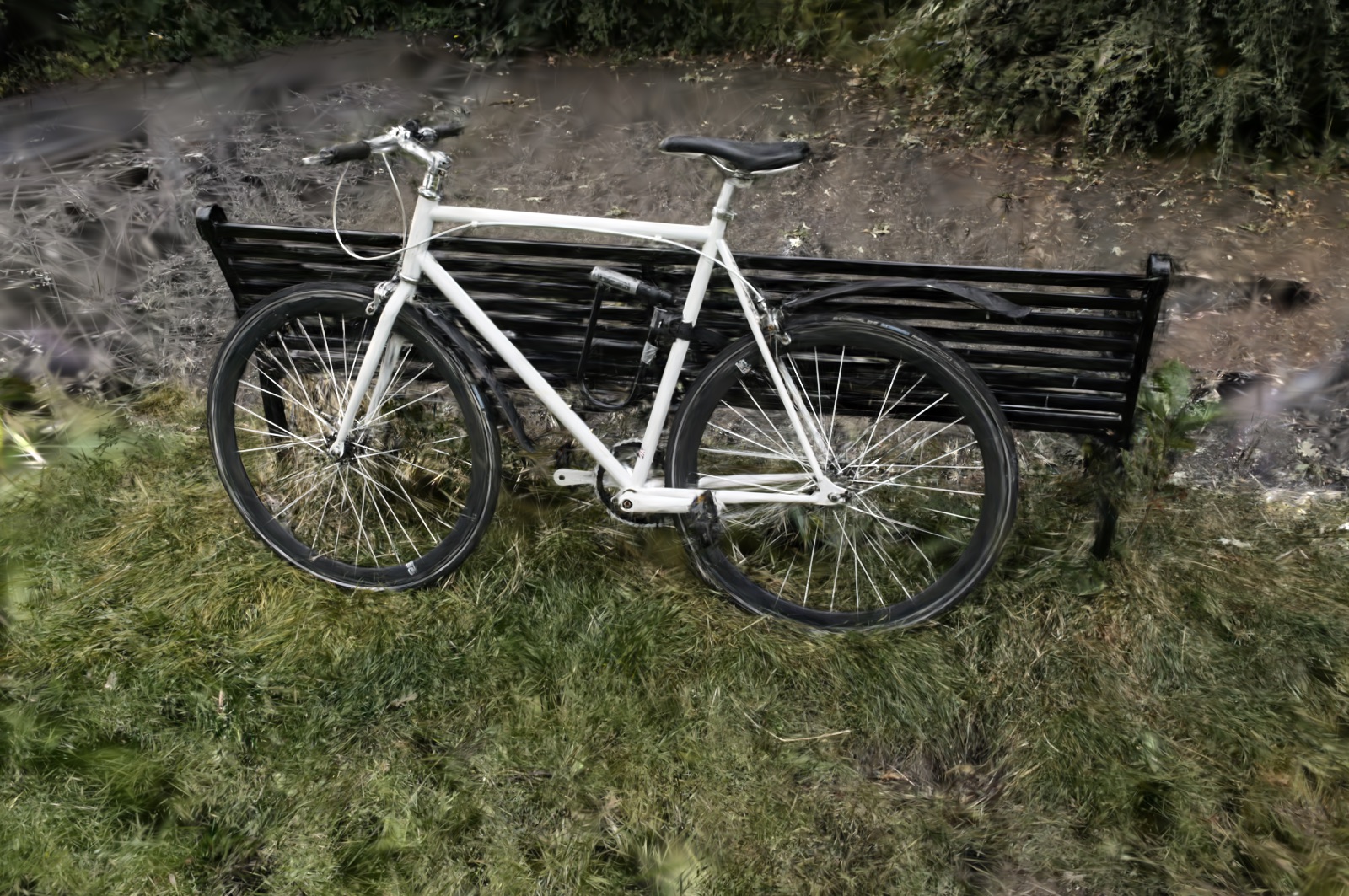}
    \end{subfigure}
    \begin{subfigure}{0.22\textwidth}
        \centering
        \includegraphics[width=\textwidth]{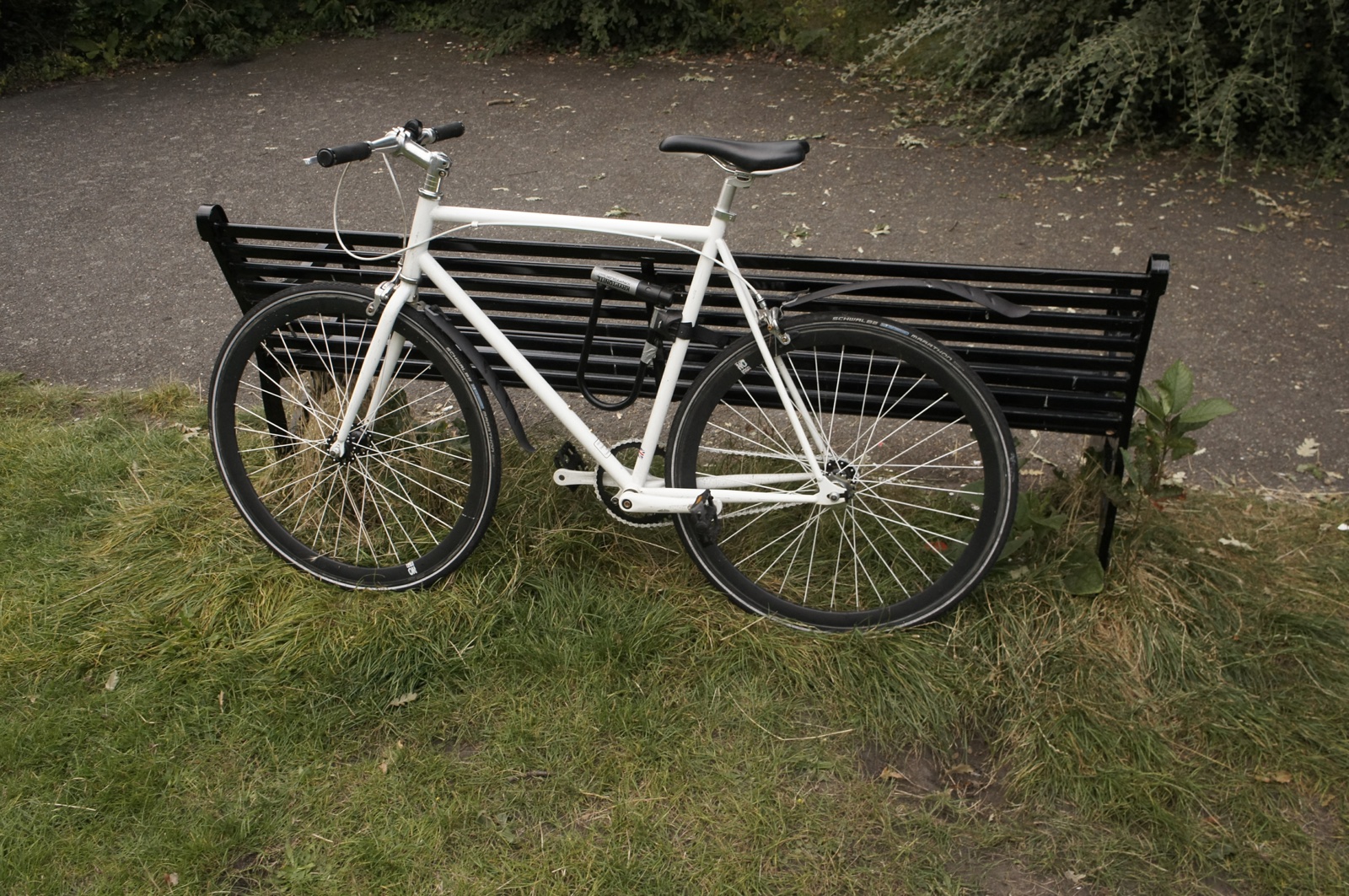}
    \end{subfigure}
    \begin{subfigure}{0.22\textwidth}
        \centering
        \includegraphics[width=\textwidth]{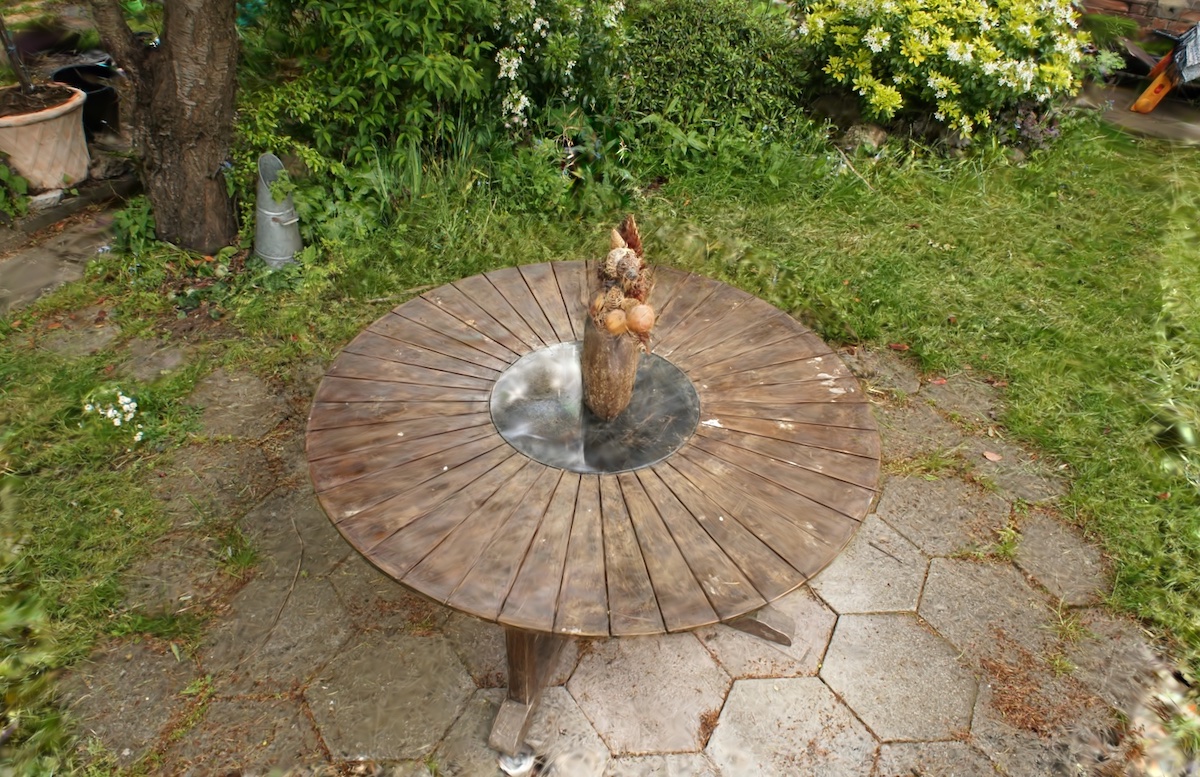}
        \caption{Uniform Sampling}
    \end{subfigure}
    \begin{subfigure}{0.22\textwidth}
        \centering
        \includegraphics[width=\textwidth]{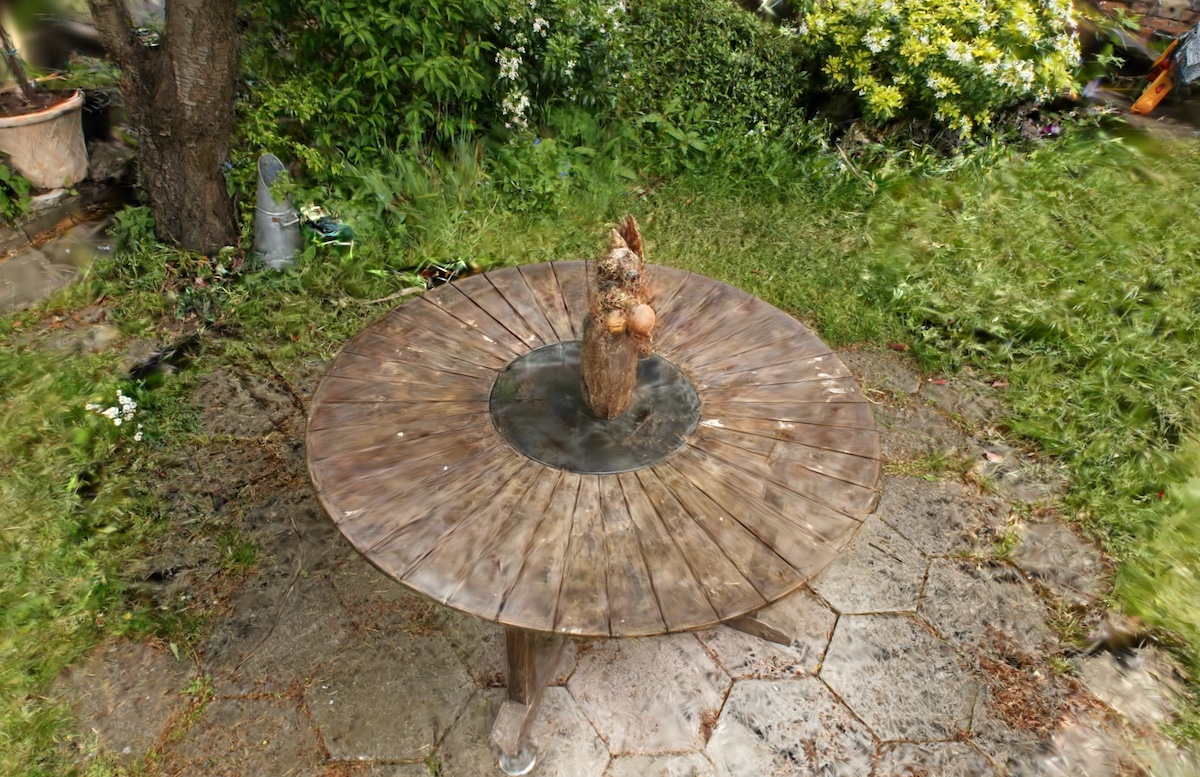}
        \caption{FisherRF}
    \end{subfigure}
    \begin{subfigure}{0.22\textwidth}
        \centering
        \includegraphics[width=\textwidth]{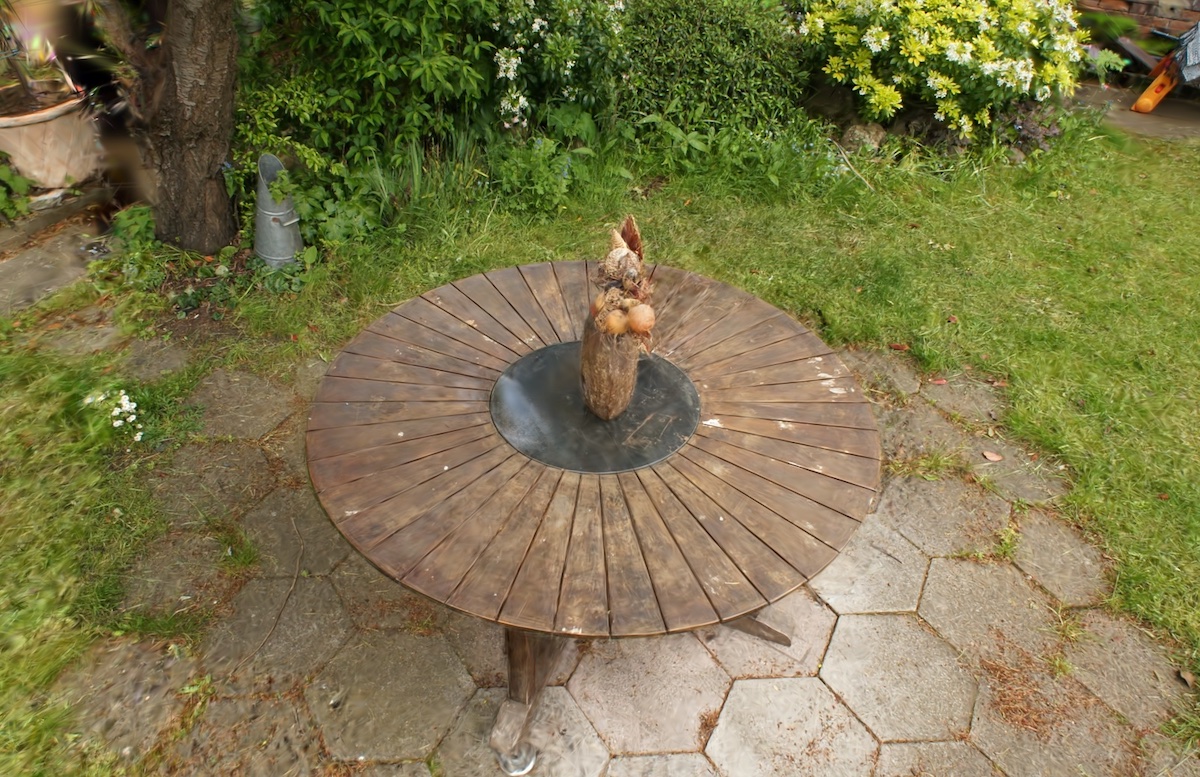}
        \caption{D-Opt. (Block)}
    \end{subfigure}
    \begin{subfigure}{0.22\textwidth}
        \centering
        \includegraphics[width=\textwidth]{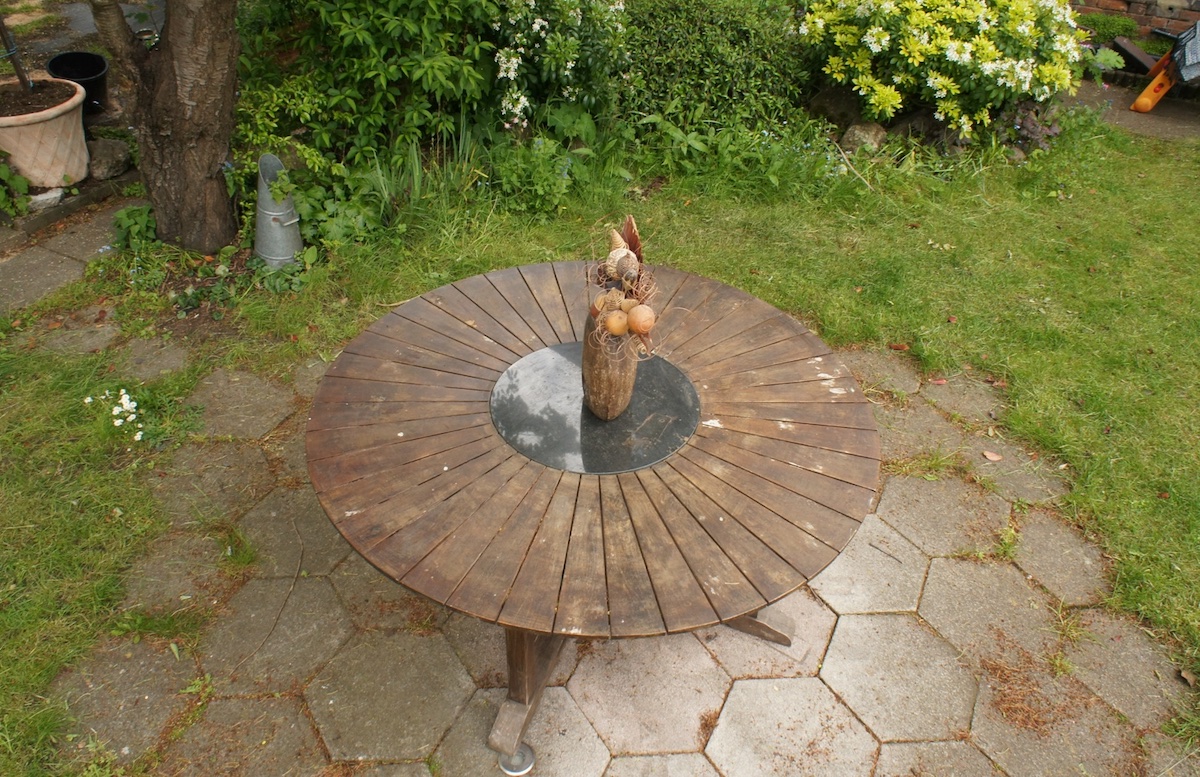}
        \caption{Ground Truth}
    \end{subfigure}
    \caption{Comparison of view selection methods on the Mip-Nerf360 dataset with 10 views. The columns are built by different methods and in order are: uniform sampling, FisherRF, Block D-GS, and the ground truth image.}
    \vspace{-4mm}
    \label{fig:10_view_mip_comparison}    
\end{figure*}

\section{Results}
In this section, we study the effectiveness of our method on quantifying information gained from obtaining new images. Following the experimental setup of FisherRF, we quantitatively and qualitatively evaluate our method against baselines on the task of single view selection, where a 3D-GS model is trained by iteratively selecting the most informative proposal view and fitting the model to a set of candidate view. Next, we compare our method against the same baselines on batch view selection. Third, we compare the ability of each method to quantify uncertainty in novel views by studying the correlation of information gain with reconstruction metrics. Last, we perform ablation studies on the parameters of the 3D-GS model to identify the most important parameters for calculating information gain. 

\textbf{Baselines:} We compare against the recently published FisherRF \cite{FisherRF}, which calculates information gain as:
\begin{equation}
        \mathcal{I}(\bm{H}_i) = \text{tr} \begin{pmatrix} (\bm{J}_i^T \bm{J}_i)  \bm{H}^{-1}_{-} \end{pmatrix} 
\end{equation}
with matrices modeled by the simple diagonal approximation. While FisherRF compared against a random and ActiveNeRF baseline \cite{ActiveNerf}, we omit these baselines as they performed worse than FisherRF in their experiments, and instead focus on comparing our results with FisherRF. We also add a uniform sampling baseline, which is slightly different from the implementation of random in FisherRF that we found was incorrectly implemented. Instead, our uniform sampling baseline samples views uniformly from the training set, which results in a high coverage of the views in the test set, especially if the training and testing views are similarly distributed. As one of the merits of our approach is a more general solution with optimal experimental design, we implement A, D, E, and T Optimality baselines for comparison. All baselines are compared over the peak signal-to-noise ratio (PSNR), structural similarity index (SSIM) \cite{SSIM}, and LPIPS metrics \cite{LPIPS}. 

\textbf{Dataset:} All methods are compared on two common radiance field datasets. First is the Mip-NeRF360 dataset \cite{MipNerf360}, which is a real-world high-resolution dataset commonly used in novel view synthesis literature as well as by FisherRF. Mip-NeRF360 contains nine scenes, with five outdoor scenes and four indoor, which we average performance over to obtain the final results. Following FisherRF and the original 3D-GS paper, we train all models at resolutions of $1060 \times 1600$ pixels. Additionally, the prior information constant is set to a value of $\lambda_\theta = 10^{-6}$ for all models. The Mip-NeRF360 dataset contains complex scenes, where the benefit of strong view selection models is clear. However, the dataset also contains some noisy images which are distributed at random, and can impact the results.

Therefore, following FisherRF, we also evaluate all models on the Blender dataset \cite{NeRF}, which contains eight high-fidelity objects modeled synthetically. While the scenes in this experiment are less complex, this dataset allows us to study information gain quantification without the stochasticity introduced by real-world noisy images.

\subsection{Single View Selection}
First, we compare all methods on single view selection, where one candidate view is selected at a time. We follow the experimental setup of FisherRF with minimal modifications, including evaluating models on the ability to select both ten views and twenty views. Note that accurate information quantification is more apparent with ten views due to the limited amount of training information. 

\begin{table}[t]
    \centering
    \caption{Results on Single View Selection with 10 Views on the Mip-Nerf360 Dataset.}
    \resizebox{0.45\textwidth}{!}{
    \begin{tabular}{l|c c c}
        Method & PSNR ($\uparrow$) & SSIM ($\uparrow$) & LPIPS ($\downarrow$) \\
        \hline
        \vspace{-2mm} \\
        Uniform Sampling & 17.29 & 0.508 & 0.432 \\
        FisherRF & 16.81 & 0.493 & 0.445\\
        \bottomrule
        A-Opt. (Simple) & 15.55 & 0.452 & 0.480\\
        E-Opt. (Simple) & 15.33 & 0.436 & 0.488 \\
        T-Opt. (Simple) & 17.91 & 0.520 & 0.420 \\
        D-Opt. (Simple) & \underline{17.95} & \underline{0.535} & \underline{0.411} \\
        \bottomrule
        D-Opt. (Block) & \textbf{18.15} & \textbf{0.548} & \textbf{0.401} \\
    \end{tabular}
    }
    \label{tab:10_view_single_mip}
    \vspace{-4mm}
\end{table}

\begin{table}[t]
    \centering
    \caption{Results on Single View Selection with 10 Views on the Blender Dataset.}
    \resizebox{0.45\textwidth}{!}{
    \begin{tabular}{l|c c c}
        Method & PSNR ($\uparrow$) & SSIM ($\uparrow$) & LPIPS ($\downarrow$) \\
        \hline
        \vspace{-2mm} \\
        Uniform Sampling & 23.32 & 0.885 & 0.101 \\
        FisherRF & 24.59 & 0.897 & 0.091 \\
        \bottomrule
        A-Opt. (Simple) & 22.39 & 0.876 & 0.116 \\
        E-Opt. (Simple) & 21.40 & 0.862 & 0.129 \\
        T-Opt. (Simple) & 25.40 & 0.908 & 0.080 \\
        D-Opt. (Simple) & \textbf{25.52} & \textbf{0.909} & \textbf{0.078}  \\
        \bottomrule
        D-Opt. (Block) & \underline{25.41} & \underline{0.908} & \underline{0.078} \\
    \end{tabular}
    }
    \vspace{-4mm}
    \label{tab:10_view_single_synthetic}
\end{table}

\textbf{Ten Views:} For the ten view setup, each method begins with 2 training views, and is trained for $100v$ iterations, where $v$ is the number of training views. The method then selects a single candidate view to add to the training set, and repeats the process until $v=10$, at which point the model trains until a cumulative total of $10,000$ training steps. Qualitative examples on the Mip-Nerf360 dataset are shown in Fig. \ref{fig:10_view_mip_comparison}, and qualitative examples on the Blender dataset are shown in Fig. \ref{fig:10_view_blender_comparison}.

First, we compare models on the Mip-NeRF360 dataset, whose performance metrics can be found in Table \ref{tab:10_view_single_mip}. In this experiment, FisherRF is slightly outperformed by the uniform sampling method. We would like to note that uniform sampling is actually an effective method for this experimental setup, and will achieve a near optimal performance if the test and train set are similarly distributed. By reframing the approach to information gain within 3D-GS as optimal experimental design, we find that both T-Optimality and D-Optimality approaches improve significantly over the FisherRF and uniform sampling baselines. Additionally, the block diagonal approximation significantly improves the structural quality measured by SSIM and LPIPS metrics.

\begin{table}[b]
    \centering
    \vspace{-4mm}
    \caption{Results on Single View Selection with 20 Views on the Mip-Nerf360 Dataset.}
    \resizebox{0.45\textwidth}{!}{
    \begin{tabular}{l|c c c}
        Method & PSNR ($\uparrow$) & SSIM ($\uparrow$) & LPIPS ($\downarrow$) \\
        \hline
        \vspace{-2mm} \\
        Uniform Sampling & 20.86 & 0.616 & 0.408 \\
        FisherRF & 20.89 & 0.608 & 0.416\\
        \bottomrule
        A-Opt. (Simple) & 18.62 & 0.558 & 0.452 \\
        E-Opt. (Simple) & 19.57 & 0.580 & 0.433 \\
        T-Opt. (Simple) & 21.07 & 0.615 & 0.409 \\
        D-Opt. (Simple) & \underline{21.09} & \underline{0.624} & \underline{0.406} \\
        \bottomrule
        D-Opt. (Block) & \textbf{21.32} & \textbf{0.636} & \textbf{0.397} \\
    \end{tabular}
    }
    \label{tab:20_view_single_mip}
\end{table}

Next, we repeat the same set of experiments on the synthetic dataset, shown in Table \ref{tab:10_view_single_synthetic}. Here, we find that FisherRF performs significantly better than the uniform sampling method, which may be due to the less complex synthetic scenes, where FisherRF is able to more accurately quantify information on a single object. Similar to before, we find that both T-Optimality and D-Optimality methods outperform the baselines by a wide margin. Across all three metrics, simple and block D-Optimality perform similarly, which we expect is due to the saturation of performance as the approaches are near optimal.


\textbf{Twenty Views:} The twenty view experimental setup follows a very similar approach, of iteratively adding views and training for $100v$ iterations before selecting the next view. However, this setup begins with $v=4$ training views, adds images until $v=20$, and trains until a cumulative total of $21,000$ training steps. Experimental results on the MIP dataset can be found in Table \ref{tab:20_view_single_mip}. Similar to the ten view experiment, we find that uniform sampling slightly outperforms FisherRF while T and D Optimality achieve the highest performance with an improvement from the block diagonal approximation.  



\subsection{Batch View Selection}

Next, we compare all methods on batch view selection, where information gain is evaluated over \textit{several} views simultaneously before training. This problem is more applicable to real-world scenarios such as information gained over a robot trajectory, or identification of the best set of views of an object.

\begin{table}[b]
    \centering
    \vspace{-4mm}
    \caption{Results on Batch View Selection on Mip-Nerf360 dataset.}
    \resizebox{0.45\textwidth}{!}{
    \begin{tabular}{l|c c c}
        Method & PSNR ($\uparrow$) & SSIM ($\uparrow$) & LPIPS ($\downarrow$) \\
        \hline
        \vspace{-2mm} \\
        Uniform Sampling & 20.42 & 0.613 & 0.389 \\
        FisherRF & 20.50 & 0.603 & 0.399\\
        \bottomrule
        A-Opt. (Simple) & 18.14 & 0.553 & 0.428 \\
        E-Opt. (Simple) & 17.88 & 0.535 & 0.440 \\
        T-Opt. (Simple) & 20.73 & 0.611 & 0.391 \\
        D-Opt. (Simple) & \textbf{20.86} & \underline{0.624} & \underline{0.383} \\
        \bottomrule
        D-Opt. (Block) & \underline{20.79} & \textbf{0.631} & \textbf{0.378} \\
    \end{tabular}
    }
    \label{tab:batch_mip}
\end{table}

\textbf{Iterative:} In the first experiment on batch view selection we follow the experimental set-up of FisherRF.  The procedure is similar to single view selection, however models begin with $4$ training views, are trained for $150v$ iterations between view selections, and select \textit{4} views at a time until $20$ views are obtained. All models are trained for a cumulative total of $10,000$ training steps. Results on the Mip-Nerf360 dataset are shown in Table \ref{tab:batch_mip}, and similar to previous experiments demonstrate superior performance of D and T optimality. Additionally, the block diagonal approximation improves structural quality measured by SSIM and LPIPS metrics. Note that due to the iterative view selection and training, this experiment may not reward batch view diversity, motivating our next experiment.

\begin{table}[t]
    \centering
    \caption{Results on Keyframe Selection on Blender Dataset.}
    \resizebox{0.45\textwidth}{!}{
    \begin{tabular}{l|c c c}
        Method & PSNR ($\uparrow$) & SSIM ($\uparrow$) & LPIPS ($\downarrow$) \\
        \hline
        \vspace{-2mm} \\
        Uniform Sampling & 23.47 & 0.888 & 0.109 \\
        FisherRF & 18.37 & 0.829 & 0.184 \\
        \bottomrule
        A-Opt. (Simple) & 17.05 & 0.811 & 0.226 \\
        E-Opt. (Simple) & 16.55 & 0.786 & 0.255 \\
        T-Opt. (Simple) & \textbf{24.90} & \textbf{0.903} & \textbf{0.096} \\
        D-Opt. (Simple) & 24.26 & 0.899 & 0.101 \\
        \bottomrule
        D-Opt. (Block) & \underline{24.53} & \underline{0.902} & \underline{0.099} \\
    \end{tabular}
    }
    \label{tab:view_compress_synthetic}
    \vspace{-4mm}
\end{table}

\textbf{Keyframe Selection:} To further study batch view selection in a setting more similar to SLAM applications, we compare each approach on keyframe selection. All methods are provided the same pre-trained 3D-GS model and select ten keyframes without replacement from a set of views. The selected views are then used to re-train a new 3D-GS model, which is evaluated and compared as a measure of batch information quantification. Results on the Blender dataset are summarized in Table \ref{tab:view_compress_synthetic}, demonstrating low performance of FisherRF, A Optimality and E Optimality which select similar views. Instead, T and D Optimality select informative and different views resulting in a large performance gap. 



\subsection{Correlation with Render Quality}
Intuitively, we expect information gain and rasterization quality at view points to be inversely related. For instance, if a 3D-GS model has only been trained on the front side of a chair, view points from the back side would have poor renderings while providing high information gain to the 3D-GS model. Therefore, as another test of our proposed method of quantifying information gain, we create a sparsification plot \cite{sparsification} to study the \textit{inverse} relationship between image information gain and image render quality. 

\begin{figure}[b]
    \centering
    \vspace{-4mm}
    \begin{subfigure}{0.49\linewidth}
        \centering
        \includegraphics[width=\textwidth]{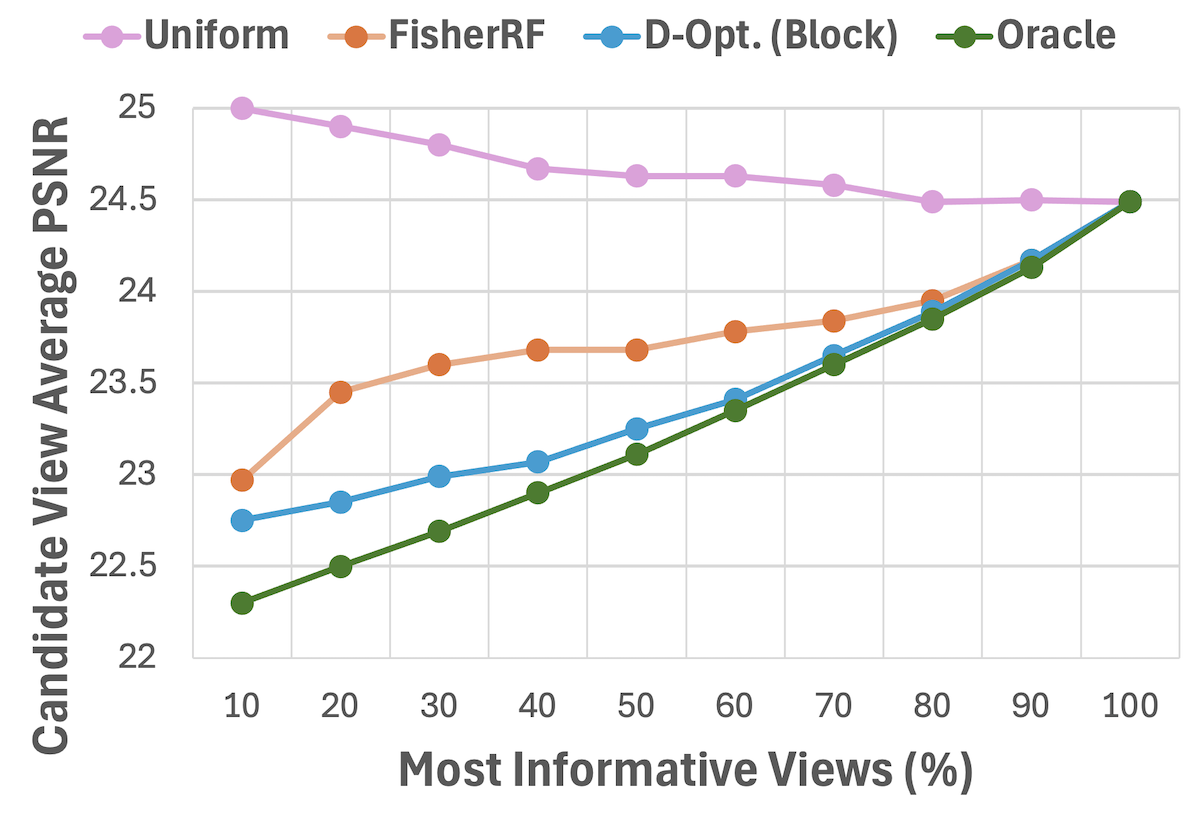}
        \caption{Ficus}
    \end{subfigure}
    \begin{subfigure}{0.49\linewidth}
        \centering
        \includegraphics[width=\textwidth]{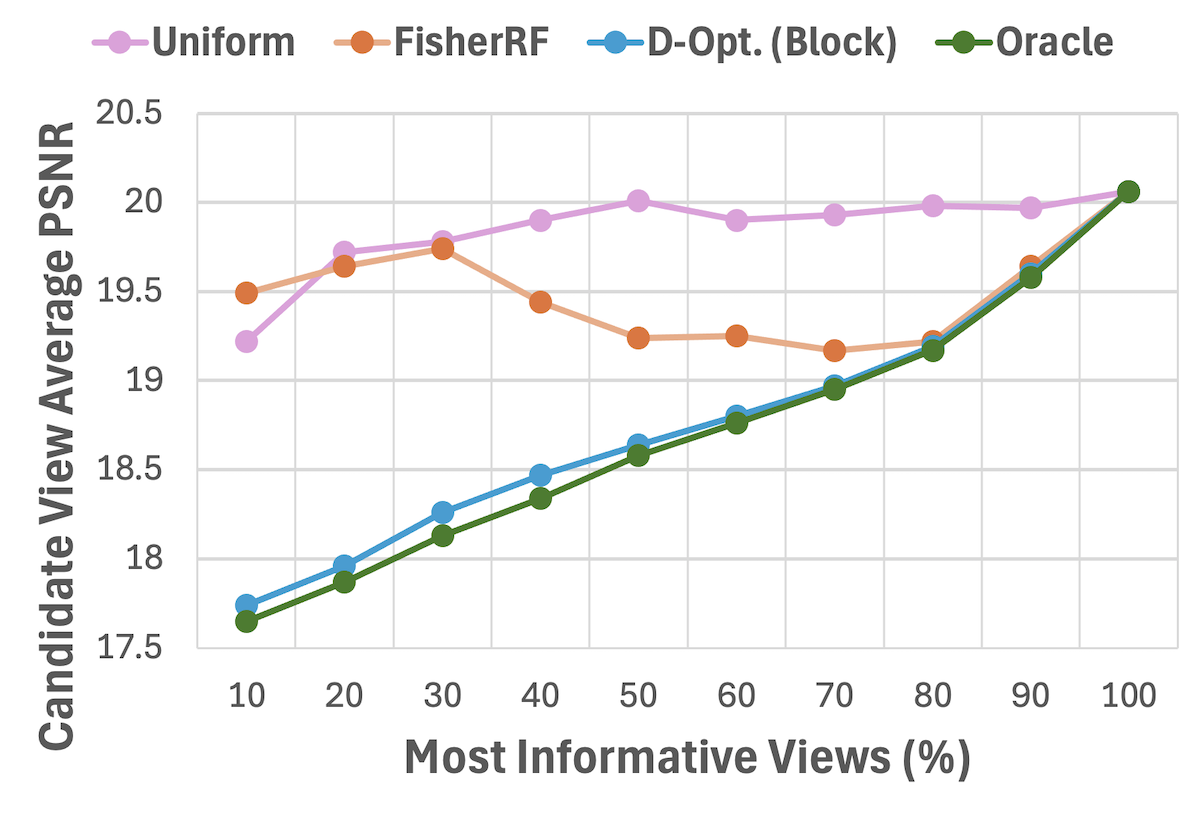}
        \caption{Drums}
    \end{subfigure}
    \caption{Correlation of expected information gain with PSNR of candidate views on two objects in Blender dataset.}
    \label{fig:information_reconstruction_correlation} 
\end{figure}

The sparsification plot in Fig. \ref{fig:information_reconstruction_correlation} is created by first training a 3D-GS model on ten randomly selected images for $2,000$ iterations. Each method, using the same random seed and trained model, sorts candidate views by expected information gain. At decile increments, the cumulative average reconstruction quality of views is calculated and plotted for each method. Reading from left to right, the plot indicates the average reconstruction quality of the most informative views. Due to the inverse nature between image uncertainty and information gain, we would expect the most informative candidate views (left) to have \textit{low} reconstruction quality. 


 We can see from Fig. \ref{fig:information_reconstruction_correlation} that D-Opt. (Block) has a monotomic relationship with the reconstruction quality, whereas FisherRF has difficulty with some objects. To better understand the relative performance, we also include plots of the Uniform Sampling and Oracle methods, where the Uniform Sampling method shows no correlation between the selected images and the reconstruction quality as expected. The Oracle baseline sorts the candidate views by the actual PSNR value, representing a perfect baseline, with similar performance to D-Opt. (Block).

\subsection{Ablation Study}
Last, we conclude with ablation studies on the most important parameters for information gain quantification. While we evaluated our methods with all parameters, reducing the number of parameters can improve computational efficiency. We compare the performance of D-Opt. (Block) with different parameter combinations in Table \ref{tab:ablation_results} on the task of single view selection with 10 views on the Blender dataset. Peculiarly, we find that the geometric parameters are important to quantifying information with few images while the spherical harmonics do not result in a significant difference, supporting the approach of PUP 3D-GS \cite{UncertaintyPruning}. We expect that the opacity decreases performance since our approach does not capture the cross-correlation of ellipsoids. Additionally, we suspect that the spherical harmonics may be more useful with a well fitted scene which already has learned the geometric structure. Nonetheless, these results indicate that information may be evaluated with a minimal set of the geometric parameters, which can increase inference speed. Concretely, on the ship scene from the Blender dataset the memory and latency are as follows: 2.62 GB at 0.15 Hz for full block diagonal, 75.3 MB at 1.71 Hz for block diagonal without spherical harmonics, and 44.4 MB at 12.16 Hz for the simple approximation.  Note that the cost of the simple approximation is the same as that of FisherRF.


\begin{table}[t]
    \centering
    \caption{Ablation study on D-Opt. (Block) parameters on Blender Dataset with 10 images. The parameters are $sh$: spherical harmonics, $\alpha$: opacity, $\mu$: location, R: rotation, and S: scale.}
    \resizebox{0.45\textwidth}{!}{
    \begin{tabular}{l|c| c c c}
        Parameters Removed & PSNR ($\uparrow$) & SSIM ($\uparrow$) & LPIPS ($\downarrow$) \\
        \hline
        \vspace{-2mm} \\
        \{$sh$ \} & 25.52 & 0.9089 & 0.0784 \\
        \{$\alpha$ \} & 25.56 & 0.9087 & 0.0778 \\ 
        \{$\mu$, R, S \} & 25.36 & 0.9074 & 0.0793 \\
        $\varnothing$ & 25.41 & 0.9084 & 0.0776 \\
    \end{tabular}
    }
    \label{tab:ablation_results}
    \vspace{-4mm}
\end{table}


\section{Conclusion}
In this paper, we introduced a novel method for calculating the information gain from images in 3D-GS which builds on prior literature of P-Optimality. Information quantification for 3D-GS is an important problem for evaluating uncertainty in novel environments, selecting key frames for SLAM algorithms, and next best view applications. Our novel formulation leads to a general solution with a simple and block diagonal information matrix approximation, with computational and performance trade-offs. We demonstrate quantitatively that formulating the information quantification with T and D Optimality improves performance compared to the state of the art, supporting results from prior literature. While our method achieves significant results quantitatively and qualitatively, the simple and block diagonal approximations discard correlation \textit{between} ellipsoids. For future work we would like to investigate re-formulating the problem to include inter-ellipsoid information, such as from the structural similarity loss or the opacity parameter.
{
\bibliographystyle{ieeenat_fullname}
\bibliography{bib/strings-abrv,bib/ieee-abrv,bib/refs}

\begin{thebibliography}{43}
\providecommand{\natexlab}[1]{#1}
\providecommand{\url}[1]{\texttt{#1}}
\expandafter\ifx\csname urlstyle\endcsname\relax
  \providecommand{\doi}[1]{doi: #1}\else
  \providecommand{\doi}{doi: \begingroup \urlstyle{rm}\Url}\fi

\bibitem[Bar-Shalom et~al.(2004)Bar-Shalom, Li, and Kirubarajan]{bar2004estimation}
Yaakov Bar-Shalom, X~Rong Li, and Thiagalingam Kirubarajan.
\newblock \emph{{Estimation with Applications to Tracking and Navigation: Theory, Algorithms and Software}}.
\newblock John Wiley \& Sons, 2004.

\bibitem[Barron et~al.(2022)Barron, Mildenhall, Verbin, Srinivasan, and Hedman]{MipNerf360}
Jonathan~T. Barron, Ben Mildenhall, Dor Verbin, Pratul~P. Srinivasan, and Peter Hedman.
\newblock { Mip-NeRF 360: Unbounded Anti-Aliased Neural Radiance Fields}.
\newblock In \emph{Proc. {IEEE} Conf. Comput. Vis. Pattern Recog.}, pages 5460--5469, 2022.

\bibitem[Carrillo et~al.(2012)Carrillo, Reid, and Castellanos]{carrillo2012onthecomparison}
Henry Carrillo, Ian Reid, and José~A. Castellanos.
\newblock {On the comparison of uncertainty criteria for active SLAM}.
\newblock In \emph{Proc. {IEEE} Int. Conf. Robot. and Automation}, pages 2080--2087, 2012.

\bibitem[Chen et~al.(2020)Chen, Huang, and Fitch]{chen2020active}
Yongbo Chen, Shoudong Huang, and Robert Fitch.
\newblock {Active SLAM for mobile robots with area coverage and obstacle avoidance}.
\newblock \emph{{IEEE/ASME} Trans. Mechatronics}, 25\penalty0 (3):\penalty0 1182--1192, 2020.

\bibitem[Cover and Thomas(2005)]{ElementsInformationTheory}
Thomas~M. Cover and Joy~A. Thomas.
\newblock \emph{{Elements of Information Theory}}, chapter~8, pages 243--259.
\newblock John Wiley \& Sons, Ltd, 2005.

\bibitem[Daxberger et~al.(2021)Daxberger, Kristiadi, Immer, Eschenhagen, Bauer, and Hennig]{LaplaceApprox}
Erik Daxberger, Agustinus Kristiadi, Alexander Immer, Runa Eschenhagen, Matthias Bauer, and Philipp Hennig.
\newblock {Laplace redux – effortless Bayesian deep learning}.
\newblock In \emph{Proc. Advances Neural Inform. Process. Syst. Conf.}, 2021.

\bibitem[Fei et~al.(2024)Fei, Xu, Zhang, Zhou, Yang, and He]{3D-GS_Survey}
Ben Fei, Jingyi Xu, Rui Zhang, Qingyuan Zhou, Weidong Yang, and Ying He.
\newblock {3D Gaussian Splatting as New Era: A Survey}.
\newblock \emph{{IEEE} Trans. Graph.}, pages 1--20, 2024.

\bibitem[Ghaffari~Jadidi et~al.(2018)Ghaffari~Jadidi, Valls~Miro, and Dissanayake]{MaaniActiveMapping}
Maani Ghaffari~Jadidi, Jaime Valls~Miro, and Gamini Dissanayake.
\newblock {Gaussian processes autonomous mapping and exploration for range-sensing mobile robots}.
\newblock \emph{Auton. Robot.}, 42\penalty0 (2):\penalty0 273--290, 2018.

\bibitem[Goli et~al.(2024)Goli, Reading, Sell{\'a}n, Jacobson, and Tagliasacchi]{BayesRays}
Lily Goli, Cody Reading, Silvia Sell{\'a}n, Alec Jacobson, and Andrea Tagliasacchi.
\newblock {Bayes' Rays: Uncertainty Quantification for Neural Radiance Fields}.
\newblock In \emph{Proc. {IEEE} Conf. Comput. Vis. Pattern Recog.}, pages 20061--20070, 2024.

\bibitem[Hanson et~al.(2024)Hanson, Tu, Singla, Jayawardhana, Zwicker, and Goldstein]{UncertaintyPruning}
Alex Hanson, Allen Tu, Vasu Singla, Mayuka Jayawardhana, Matthias Zwicker, and Tom Goldstein.
\newblock {PUP 3D-GS: Principled Uncertainty Pruning for 3D Gaussian Splatting}.
\newblock \emph{arXiv}, abs/2406.10219, 2024.

\bibitem[Huan and Marzouk(2013)]{Huan2013b}
Xun Huan and Youssef~M. Marzouk.
\newblock {Simulation-based optimal Bayesian experimental design for nonlinear systems}.
\newblock \emph{J. of Comput. Phys.}, 232\penalty0 (1):\penalty0 288--317, 2013.

\bibitem[Huan and Marzouk(2014)]{Huan2014a}
Xun Huan and Youssef~M. Marzouk.
\newblock {Gradient-based stochastic optimization methods in Bayesian experimental design}.
\newblock \emph{Int. J. for Uncertainty Quant.}, 4\penalty0 (6):\penalty0 479--510, 2014.

\bibitem[Ilg et~al.(2018)Ilg, {\c{C}}i{\c{c}}ek, Galesso, Klein, Makansi, Hutter, and Brox]{sparsification}
Eddy Ilg, {\"O}zg{\"u}n {\c{C}}i{\c{c}}ek, Silvio Galesso, Aaron Klein, Osama Makansi, Frank Hutter, and Thomas Brox.
\newblock {Uncertainty Estimates and Multi-hypotheses Networks for Optical Flow}.
\newblock In \emph{Proc. European Conf. Comput. Vis.}, pages 677--693, 2018.

\bibitem[Jiang et~al.(2024)Jiang, Lei, and Daniilidis]{FisherRF}
Wen Jiang, Boshu Lei, and Kostas Daniilidis.
\newblock {FisherRF: Active View Selection and Uncertainty Quantification for Radiance Fields using Fisher Information}.
\newblock In \emph{Proc. European Conf. Comput. Vis.}, pages 422--440, 2024.

\bibitem[Kerbl et~al.(2023)Kerbl, Kopanas, Leimk{\"u}hler, and Drettakis]{3D-GS}
Bernhard Kerbl, Georgios Kopanas, Thomas Leimk{\"u}hler, and George Drettakis.
\newblock {3D Gaussian Splatting for Real-Time Radiance Field Rendering}.
\newblock \emph{{IEEE} Trans. Graph.}, 42\penalty0 (4), 2023.

\bibitem[Kiefer(1974)]{kiefer1974general}
Jack Kiefer.
\newblock {General equivalence theory for optimum designs (approximate theory)}.
\newblock \emph{{The annals of Statistics}}, pages 849--879, 1974.

\bibitem[Kirsch and Gal(2022)]{FisherInformation}
Andreas Kirsch and Yarin Gal.
\newblock {Unifying Approaches in Active Learning and Active Sampling via Fisher Information and Information-Theoretic Quantities}.
\newblock \emph{J. Mach. Learning Res.}, 2022.

\bibitem[Kullback and Leibler(1951)]{KLD}
S. Kullback and R.~A. Leibler.
\newblock {On Information and Sufficiency}.
\newblock \emph{The Ann. Math. Stat.}, 22\penalty0 (1):\penalty0 79 -- 86, 1951.

\bibitem[Li and ming Cheung(2024)]{Variational3DGS}
Ruiqi Li and Yiu ming Cheung.
\newblock {Variational Multi-scale Representation for Estimating Uncertainty in 3D Gaussian Splatting}.
\newblock In \emph{Proc. Advances Neural Inform. Process. Syst. Conf.}, 2024.

\bibitem[Lindley(1956{\natexlab{a}})]{ExperimentInformation}
D.~V. Lindley.
\newblock {On a Measure of the Information Provided by an Experiment}.
\newblock \emph{The Ann. Math. Stat.}, 27\penalty0 (4):\penalty0 986 -- 1005, 1956{\natexlab{a}}.

\bibitem[Lindley(1956{\natexlab{b}})]{MeasureInformation}
D.~V. Lindley.
\newblock {On a Measure of the Information Provided by an Experiment}.
\newblock \emph{The Ann. Math. Stat.}, 27\penalty0 (4):\penalty0 986--1005, 1956{\natexlab{b}}.

\bibitem[Liu et~al.(2024)Liu, Jiang, Lei, Pandey, Daniilidis, and Motee]{FisherRFPerception}
Guangyi Liu, Wen Jiang, Boshu Lei, Vivek Pandey, Kostas Daniilidis, and Nader Motee.
\newblock {Beyond Uncertainty: Risk-Aware Active View Acquisition for Safe Robot Navigation and 3D Scene Understanding with FisherRF}.
\newblock \emph{arXiv}, abs/2403.11396, 2024.

\bibitem[Martinez-Cantin et~al.(2009)Martinez-Cantin, de~Freitas, Brochu, Castellanos, and Doucet]{ActiveSLAM}
Ruben Martinez-Cantin, Nando de Freitas, Eric Brochu, Jos{\'e}~A. Castellanos, and A. Doucet.
\newblock A bayesian exploration-exploitation approach for optimal online sensing and planning with a visually guided mobile robot.
\newblock \emph{Auton. Robot.}, 27:\penalty0 93--103, 2009.

\bibitem[Matsuki et~al.(2024)Matsuki, Murai, Kelly, and Davison]{3D-GS_SLAM}
Hidenobu Matsuki, Riku Murai, Paul~H.J. Kelly, and Andrew~J. Davison.
\newblock {Gaussian Splatting SLAM}.
\newblock In \emph{Proc. {IEEE} Conf. Comput. Vis. Pattern Recog.}, pages 18039--18048, 2024.

\bibitem[Mihaylova et~al.(2003)Mihaylova, Lefebvre, Bruyninckx, Gadeyne, and De~Schutter]{mihaylova2003comparison}
Lyudmila Mihaylova, Tine Lefebvre, Herman Bruyninckx, Klaas Gadeyne, and Joris De~Schutter.
\newblock {A comparison of decision making criteria and optimization methods for active robotic sensing}.
\newblock In \emph{{Numerical Methods and Applications}}, pages 316--324. Springer, 2003.

\bibitem[Mildenhall et~al.(2020)Mildenhall, Srinivasan, Tancik, Barron, Ramamoorthi, and Ng]{NeRF}
Ben Mildenhall, Pratul~P. Srinivasan, Matthew Tancik, Jonathan~T. Barron, Ravi Ramamoorthi, and Ren Ng.
\newblock {NeRF: Representing Scenes as Neural Radiance Fields for View Synthesis}.
\newblock In \emph{Proc. European Conf. Comput. Vis.}, pages 405--421, 2020.

\bibitem[Pan et~al.(2022)Pan, Lai, Song, and Huang]{ActiveNerf}
Xuran Pan, Zihang Lai, Shiji Song, and Gao Huang.
\newblock {ActiveNeRF: Learning Where to See with Uncertainty Estimation}.
\newblock In \emph{Proc. European Conf. Comput. Vis.}, pages 230--246, 2022.

\bibitem[Placed and Castellanos(2022)]{placed2022general}
Julio~A Placed and Jos{\'e}~A Castellanos.
\newblock {A General Relationship between Optimality Criteria and Connectivity Indices for Active Graph-SLAM}.
\newblock \emph{{IEEE} Robot. Autom. Letter.}, 8\penalty0 (2):\penalty0 816--823, 2022.

\bibitem[Placed et~al.(2022{\natexlab{a}})Placed, Rodr{\'\i}guez, Tard{\'o}s, and Castellanos]{placed2022explorb}
Julio~A Placed, Juan J~G{\'o}mez Rodr{\'\i}guez, Juan~D Tard{\'o}s, and Jos{\'e}~A Castellanos.
\newblock Explorb-slam: Active visual slam exploiting the pose-graph topology.
\newblock In \emph{Iberian Robotics conference}, pages 199--210. Springer, 2022{\natexlab{a}}.

\bibitem[Placed et~al.(2022{\natexlab{b}})Placed, Strader, Carrillo, Atanasov, Indelman, Carlone, and Castellanos]{placed2023surveyactivesimultaneouslocalization}
Julio~A. Placed, Jared Strader, Henry Carrillo, Nikolay Atanasov, Vadim Indelman, Luca Carlone, and José~A. Castellanos.
\newblock {A Survey on Active Simultaneous Localization and Mapping: State of the Art and New Frontiers}.
\newblock \emph{{IEEE} Trans. Robot.}, 39:\penalty0 1686--1705, 2022{\natexlab{b}}.

\bibitem[Qin et~al.(2024)Qin, Li, Zhou, Wang, and Pfister]{LangSplat}
Minghan Qin, Wanhua Li, Jiawei Zhou, Haoqian Wang, and Hanspeter Pfister.
\newblock {LangSplat: 3D Language Gaussian Splatting}.
\newblock In \emph{Proc. {IEEE} Conf. Comput. Vis. Pattern Recog.}, pages 20051--20060, 2024.

\bibitem[Rodríguez-Arévalo et~al.(2018)Rodríguez-Arévalo, Neira, and Castellanos]{rodriguez2018ontheimportance}
María~L. Rodríguez-Arévalo, José Neira, and José~A. Castellanos.
\newblock {On the Importance of Uncertainty Representation in Active SLAM}.
\newblock \emph{{IEEE} Trans. Robot.}, 34\penalty0 (3):\penalty0 829--834, 2018.

\bibitem[Sim and Roy(2005)]{sim2005global}
Robert Sim and Nicholas Roy.
\newblock {Global a-optimal robot exploration in slam}.
\newblock In \emph{Proc. {IEEE} Int. Conf. Robot. and Automation}, pages 661--666. IEEE, 2005.

\bibitem[Strong et~al.(2024)Strong, Lei, Swann, Jiang, Daniilidis, and au2]{FisherRFManipulation}
Matthew Strong, Boshu Lei, Aiden Swann, Wen Jiang, Kostas Daniilidis, and Monroe Kennedy~III au2.
\newblock {Next Best Sense: Guiding Vision and Touch with FisherRF for 3D Gaussian Splatting}.
\newblock \emph{arXiv}, abs/2410.04680, 2024.

\bibitem[Tancik et~al.(2022)Tancik, Casser, Yan, Pradhan, Mildenhall, Srinivasan, Barron, and Kretzschmar]{BlockNerf}
Matthew Tancik, Vincent Casser, Xinchen Yan, Sabeek Pradhan, Ben Mildenhall, Pratul~P. Srinivasan, Jonathan~T. Barron, and Henrik Kretzschmar.
\newblock {Block-NeRF: Scalable Large Scene Neural View Synthesis}.
\newblock In \emph{Proc. {IEEE} Conf. Comput. Vis. Pattern Recog.}, pages 8248--8258, 2022.

\bibitem[Vaart(1998)]{AsymptoticStatistics}
A.~W. van~der Vaart.
\newblock \emph{{Asymptotic Statistics}}, chapter~4, page 35–40.
\newblock Cambridge University Press, 1998.

\bibitem[Wang et~al.(2004)Wang, Bovik, Sheikh, and Simoncelli]{SSIM}
Zhou Wang, A.C. Bovik, H.R. Sheikh, and E.P. Simoncelli.
\newblock {Image quality assessment: from error visibility to structural similarity}.
\newblock \emph{{IEEE} Trans. Image Process.}, 13\penalty0 (4):\penalty0 600--612, 2004.

\bibitem[Wilson et~al.(2024{\natexlab{a}})Wilson, Almeida, Sun, Mahajan, Ghaffari, Ewen, Ghasemalizadeh, Kuo, and Sen]{ContinuousSemanticSplatting}
Joey Wilson, Marcelino Almeida, Min Sun, Sachit Mahajan, Maani Ghaffari, Parker Ewen, Omid Ghasemalizadeh, Cheng-Hao Kuo, and Arnie Sen.
\newblock {Modeling Uncertainty in 3D Gaussian Splatting through Continuous Semantic Splatting}.
\newblock \emph{ArXiv}, abs/2411.02547, 2024{\natexlab{a}}.

\bibitem[Wilson et~al.(2024{\natexlab{b}})Wilson, Fu, Friesen, Ewen, Capodieci, Jayakumar, Barton, and Ghaffari]{ConvBKIJournal}
Joey Wilson, Yuewei Fu, Joshua Friesen, Parker Ewen, Andrew Capodieci, Paramsothy Jayakumar, Kira Barton, and Maani Ghaffari.
\newblock {ConvBKI: Real-Time Probabilistic Semantic Mapping Network With Quantifiable Uncertainty}.
\newblock \emph{{IEEE} Trans. Robot.}, 40:\penalty0 4648--4667, 2024{\natexlab{b}}.

\bibitem[Yan et~al.(2024)Yan, Qu, Xu, Zhao, Wang, Wang, and Li]{GS-SLAM}
Chi Yan, Delin Qu, Dan Xu, Bin Zhao, Zhigang Wang, Dong Wang, and Xuelong Li.
\newblock {GS-SLAM: Dense Visual SLAM with 3D Gaussian Splatting}.
\newblock In \emph{Proc. {IEEE} Conf. Comput. Vis. Pattern Recog.}, pages 19595--19604, 2024.

\bibitem[Zhang et~al.(2018)Zhang, Isola, Efros, Shechtman, and Wang]{LPIPS}
Richard Zhang, Phillip Isola, Alexei~A. Efros, Eli Shechtman, and Oliver Wang.
\newblock { The Unreasonable Effectiveness of Deep Features as a Perceptual Metric}.
\newblock In \emph{Proc. {IEEE} Conf. Comput. Vis. Pattern Recog.}, pages 586--595, 2018.

\bibitem[Zhou et~al.(2024)Zhou, Chang, Jiang, Fan, Zhu, Xu, Chari, You, Wang, and Kadambi]{Feature3D-GS}
Shijie Zhou, Haoran Chang, Sicheng Jiang, Zhiwen Fan, Zehao Zhu, Dejia Xu, Pradyumna Chari, Suya You, Zhangyang Wang, and Achuta Kadambi.
\newblock {Feature 3dgs: Supercharging 3d gaussian splatting to enable distilled feature fields}.
\newblock In \emph{Proc. {IEEE} Conf. Comput. Vis. Pattern Recog.}, pages 21676--21685, 2024.

\bibitem[Zhu et~al.(2024)Zhu, Li, Sandström, Huang, Schindler, and Armeni]{3D-GS_SLAM2}
Liyuan Zhu, Yue Li, Erik Sandström, Shengyu Huang, Konrad Schindler, and Iro Armeni.
\newblock {LoopSplat: Loop Closure by Registering 3D Gaussian Splats}.
\newblock \emph{arXiv}, abs/2408.10154, 2024.

\end{thebibliography}
}

\clearpage
\setcounter{page}{1}
\maketitlesupplementary

\section{Twenty View Blender Results} \label{sec:rationale}
Due to page constraints, we were unable to report the results of each method on twenty view selection of the Blender dataset with the single or batch view schemes. In the main paper, we focused on ten views with the Blender dataset, as we find that the results saturate when more views are added. 

Single view selection results are shown in Table \ref{tab:20_view_single_synthetic}, where FisherRF, T and D optimality achieve significantly higher performance than uniform sampling. T and D optimality outperform FisherRF, and have similar saturated results to each other.

Results on the batch view selection are demonstrated in Table \ref{tab:batch_synthetic}, where we find similar results as the twenty view experiment. FisherRF, T and D optimality outperform the other baselines with T and D optimality achieving the highest results. Due to the large number of views and simple scenes, results are saturated between T and D optimality.

\begin{table}[t]
    \centering
    \caption{Results on Single View Selection with 20 Views on the Blender Dataset.}
    \resizebox{0.48\textwidth}{!}{
    \begin{tabular}{l|c c c}
        Method & PSNR ($\uparrow$) & SSIM ($\uparrow$) & LPIPS ($\downarrow$) \\
        \hline
        \vspace{-2mm} \\
        Uniform Sampling & 26.15 & 0.918 & 0.084 \\
        FisherRF & 27.12 & 0.925 & 0.079\\
        \bottomrule 
        A-Opt. (Simple) & 24.88 & 0.908 & 0.094 \\
        E-Opt. (Simple) & 24.87 & 0.901 & 0.097 \\
        T-Opt. (Simple) & \textbf{27.29} & 0.929 & 0.076 \\
        D-Opt. (Simple) & 27.25 & \textbf{0.930} &  \textbf{0.075} \\
        \bottomrule
        D-Opt. (Block) & \underline{27.28} & \textbf{0.930} & \textbf{0.075} \\
    \end{tabular}
    }
    \label{tab:20_view_single_synthetic}
\end{table}

\begin{table}[t]
    \centering
    \caption{Results on Batch View Selection on Blender dataset.}
    \resizebox{0.48\textwidth}{!}{
    \begin{tabular}{l|c c c}
        Method & PSNR ($\uparrow$) & SSIM ($\uparrow$) & LPIPS ($\downarrow$) \\
        \hline
        \vspace{-2mm} \\
        Uniform Sampling & 26.64 & 0.925 & 0.074 \\
        FisherRF & 27.64 & 0.932 & 0.069 \\
        \bottomrule
        A-Opt. (Simple) & 25.61 & 0.916 & 0.082 \\
        E-Opt. (Simple) & 24.99 & 0.908 & 0.087 \\
        T-Opt. (Simple) & \textbf{27.89} & \underline{0.936} & \underline{0.065} \\
        D-Opt. (Simple) & \underline{27.87} & \textbf{0.937} & \textbf{0.064} \\
        \bottomrule
        D-Opt. (Block) & 27.80 & 0.935 & \underline{0.065} \\
    \end{tabular}
    }
    \label{tab:batch_synthetic}
\end{table}

\section{Keyframe Selection} \label{sec:compression}
Next we present results from the keyframe selection experiment on the Mip-Nerf360 dataset. In this experiment, we find a large performance gap between FisherRF, A and E Optimality and the other methods similar to with the Blender dataset. T and D optimality outperform the uniform sampling baseline, however results are saturated with ten well-chosen views. Despite the performance saturation, the block diagonal approximation leads to a noticeable improvement in SSIM and LPIPS metrics. 

\begin{table}[t]
    \centering
    \caption{Results on Keyframe Selection on Mip-Nerf360 Dataset.}
    \resizebox{0.48\textwidth}{!}{
    \begin{tabular}{l|c c c}
        Method & PSNR ($\uparrow$) & SSIM ($\uparrow$) & LPIPS ($\downarrow$) \\
        \hline
        \vspace{-2mm} \\
        Uniform Sampling & 18.30 & 0.560 & 0.435 \\
        FisherRF & 15.66 & 0.471 & 0.515 \\
        \bottomrule
        A-Opt. (Simple) & 15.67 & 0.479 & 0.519 \\
        E-Opt. (Simple) & 15.96 & 0.475 & 0.524 \\
        T-Opt. (Simple) & \underline{18.66} & \underline{0.560} & \underline{0.425} \\
        D-Opt. (Simple) & 18.57 & 0.559 & 0.426 \\
        \bottomrule
        D-Opt. (Block) & \textbf{18.73} & \textbf{0.571} & \textbf{0.417} \\
    \end{tabular}
    }
    \label{tab:view_compress_real}
\end{table}

\begin{figure*}[t]
    \centering
    \begin{subfigure}{0.33\textwidth}
        \centering
        \includegraphics[width=\textwidth]{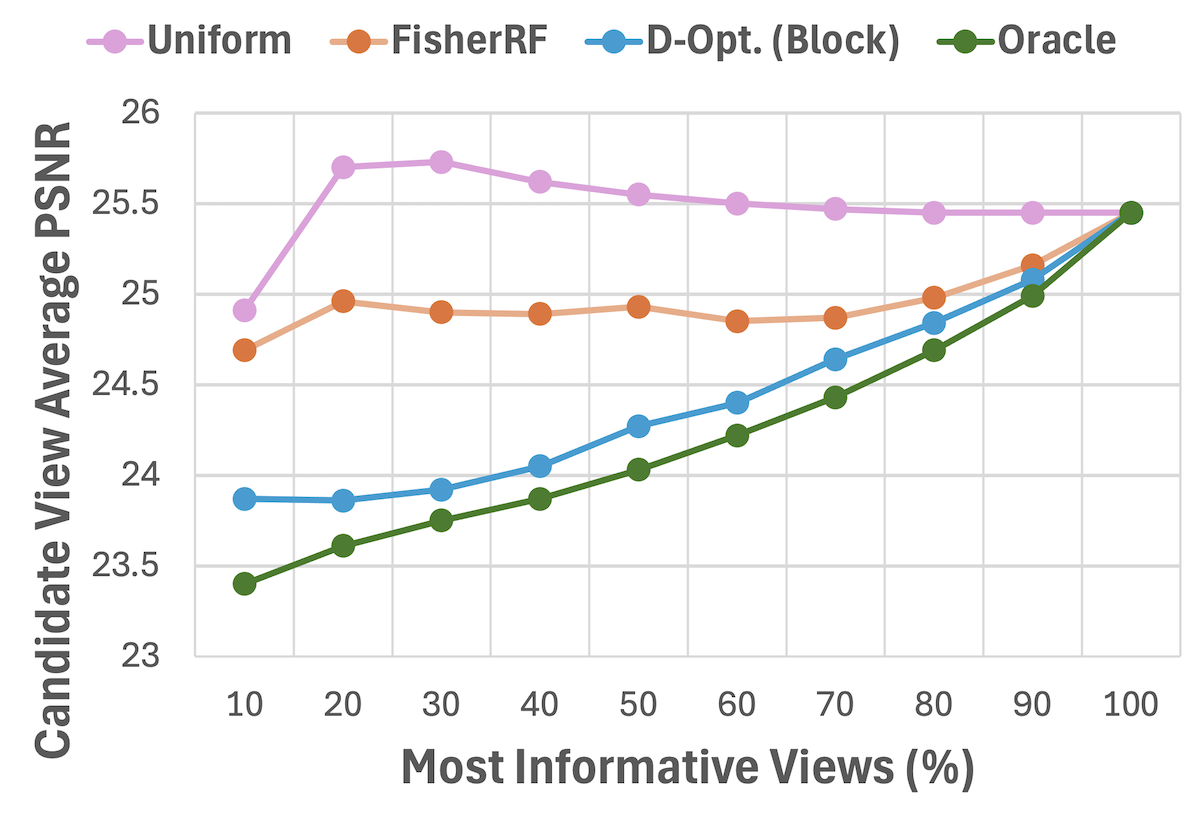}
        \caption{Chair}
    \end{subfigure}
    \begin{subfigure}{0.33\textwidth}
        \centering
        \includegraphics[width=\textwidth]{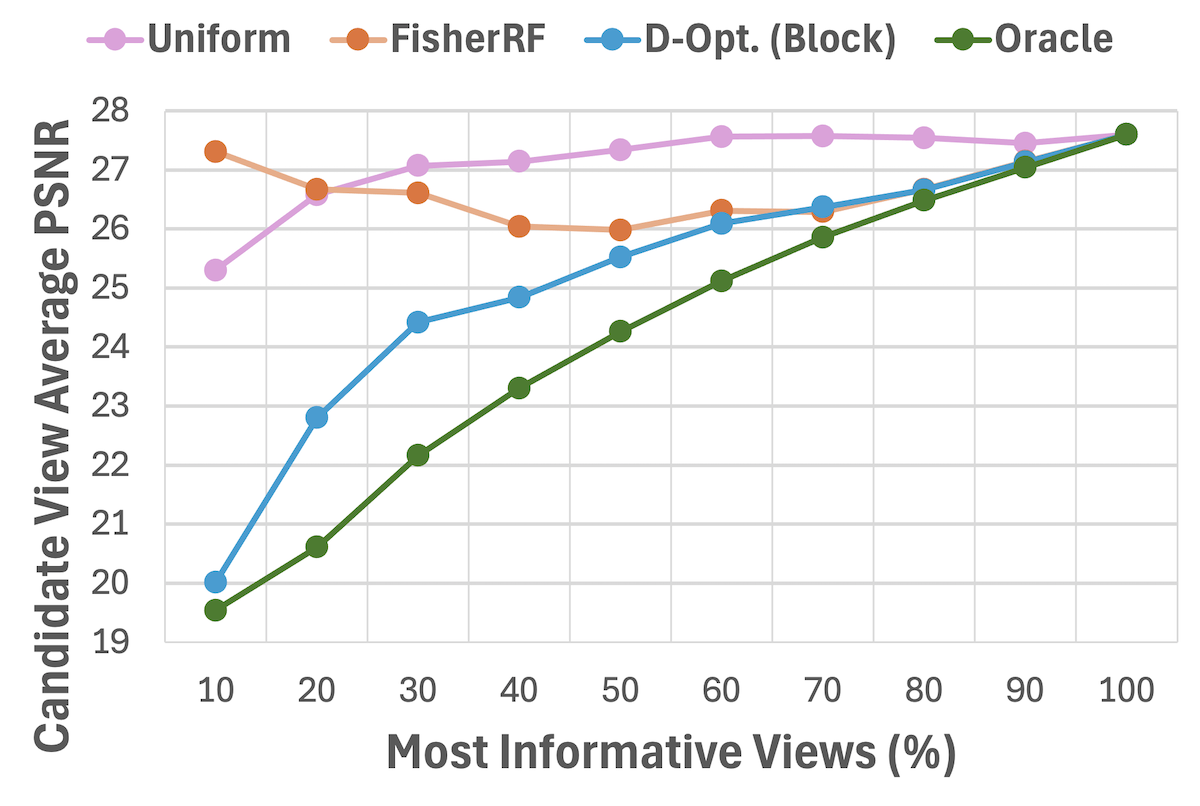}
        \caption{Hotdog}
    \end{subfigure}
    \begin{subfigure}{0.33\textwidth}
        \centering
        \includegraphics[width=\textwidth]{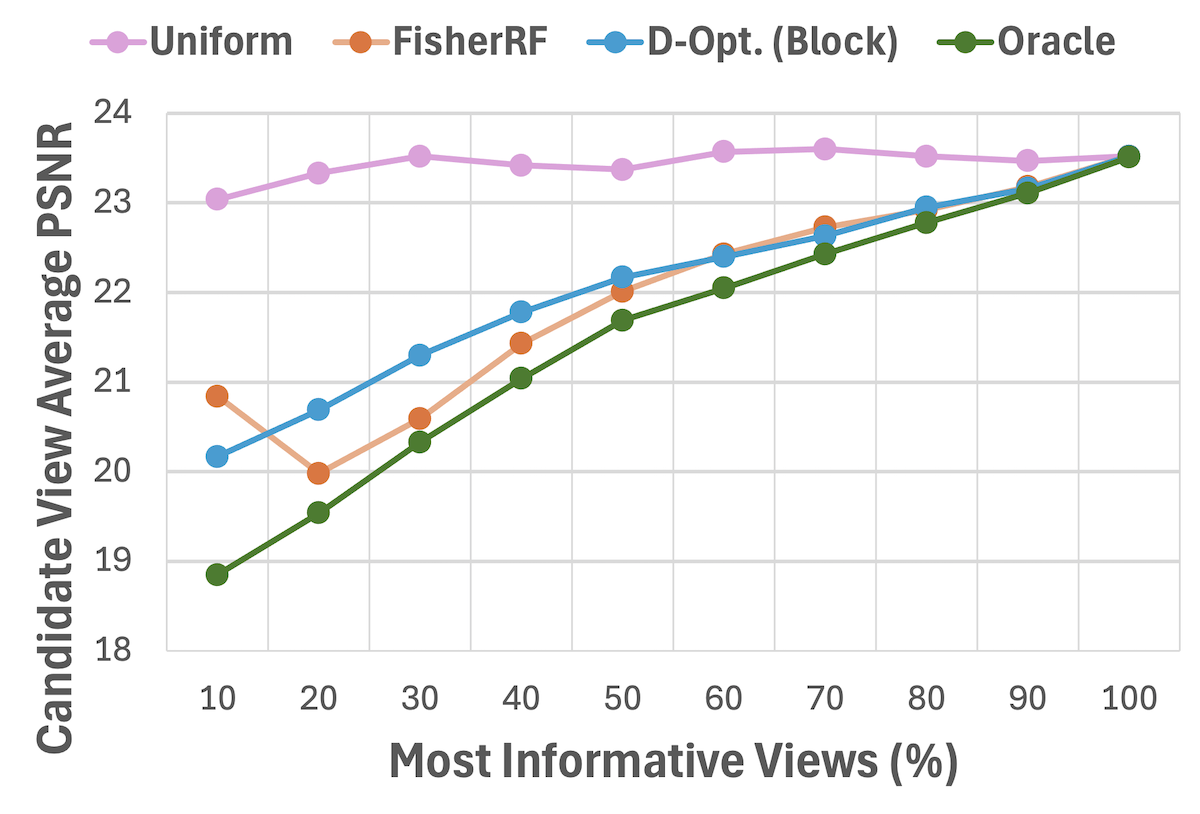}
        \caption{Lego}
    \end{subfigure}
    \begin{subfigure}{0.33\textwidth}
        \centering
        \includegraphics[width=\textwidth]{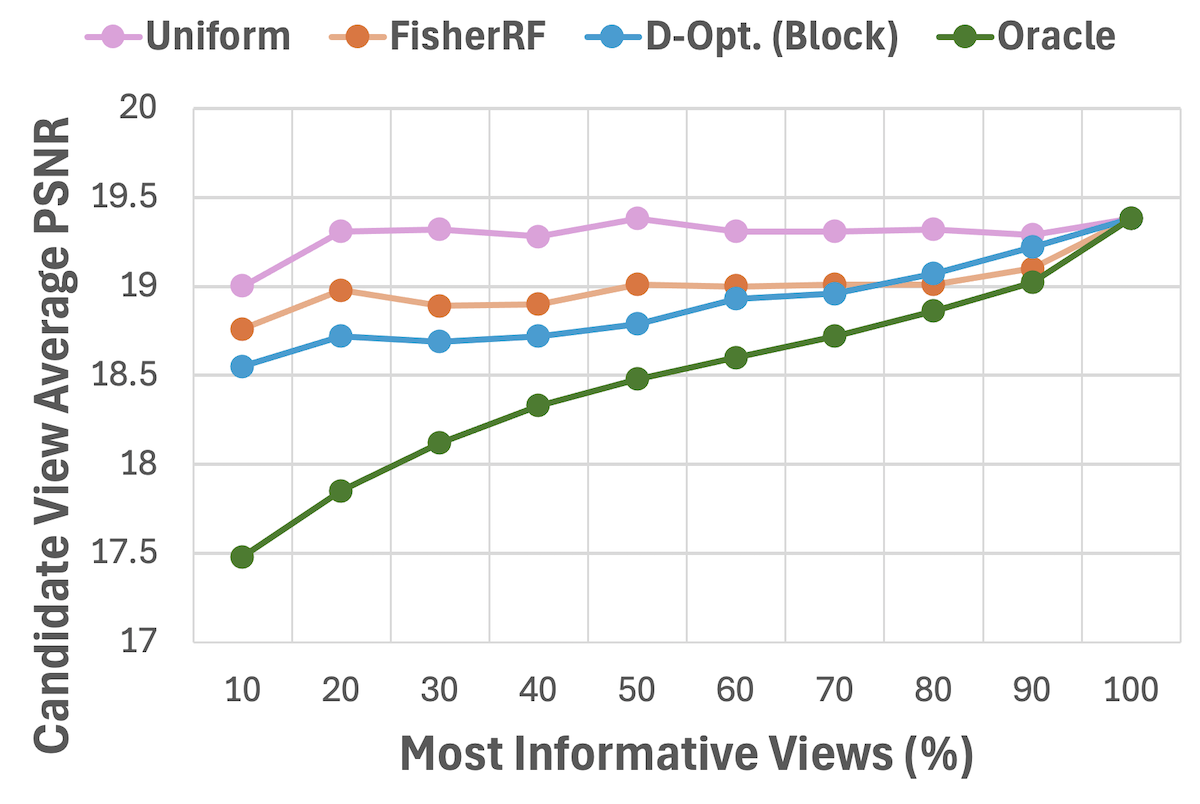}
        \caption{Materials}
    \end{subfigure}
    \begin{subfigure}{0.33\textwidth}
        \centering
        \includegraphics[width=\textwidth]{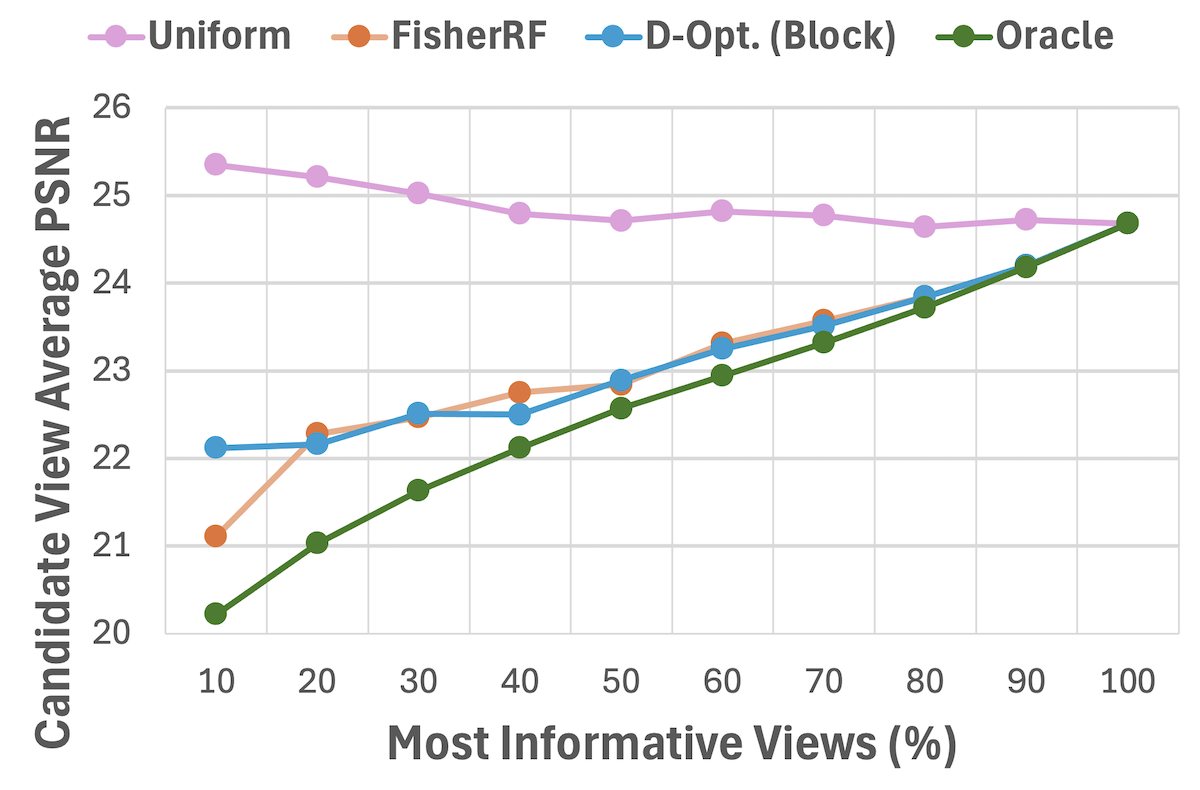}
        \caption{Mic}
    \end{subfigure}
    \begin{subfigure}{0.33\textwidth}
        \centering
        \includegraphics[width=\textwidth]{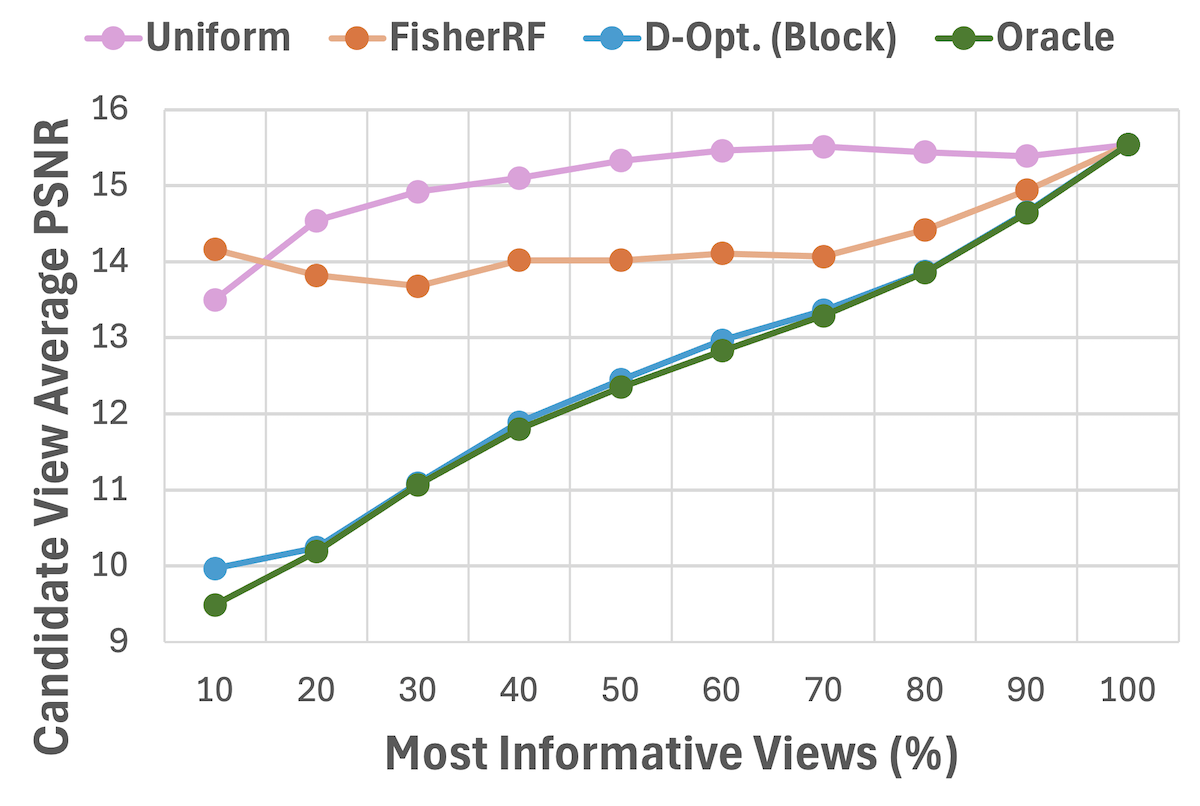}
        \caption{Ship}
    \end{subfigure}
    \caption{Uncertainty correlation plots on all remaining scenes of the Blender dataset. The oracle represents a perfect sorting of the views by PSNR. If the information gained by candidate views is well calibrated, the ordering should be similar to that of the oracle, resulting in a low value at the left of the plot which contains the average reconstruction quality of the most informative views.}
    \label{fig:blender_comparison_correlation}    
\end{figure*}

\section{Render Quality Correlation on All Scenes} \label{sec:correlation}
In this section, we first provide a more detailed explanation of the experimental setup for our study of the correlation between information gain and render quality. Next, we provide the sparsification plots for the remaining objects in Fig. \ref{fig:blender_comparison_correlation}.

In addition to identifying the most important training images, another key problem for applying 3D-GS to real-world applications is uncertainty quantification. Since information gain is dependent on the amount of information already present in a trained 3D-GS model at a candidate view, we would expect an inverse relationship between information gain and render quality. Therefore, we leverage sparsification plots to study the correlation between information gain and render quality. Intuitively, if a 3D-GS model has already observed data similar to a view, the method should quantify small information gain and the render should have a high reconstruction quality. Similarly, a candidate view with high information gain implies the model has limited information on the viewpoint, and the render would likely have poor reconstruction quality. This also follows from the inverse relationship between uncertainty and information stated in Section \ref{sec:uncertainty_relationship}.

In order to study this relationship, we leverage sparsification plots. The primary idea behind sparsification plots is to sort candidate views by information gain, and observe the relationship between information gain at candidate views and render quality at the candidate views. In this experiment, we train a single 3D-GS model on ten randomly chosen views so that there is new information at the remaining views in the dataset. Next, each method sorts the remaining views by estimated information gain from most information to least information. The purpose of the sparsification plot is then to study how the views are sorted by estimated information gain. 

To create the sparsification plot, the sorted views are organized into groups based on decile. For example, for one hundred candidate views the first group would contain the ten most informative views and the final group would contain the ten least informative views. Next, the groups are combined iteratively beginning from the most informative views and the average PSNR is calculated for the combined groups. Therefore, the plot represents the \textit{average} PSNR of the $x \%$ most informative views. As a result, we would expect to see a low PSNR for the most informative views at the left of the plot, and all methods converge at the right of the plot when calculating the average PSNR over all images.

For baselines we use the uniform sampling method, which should demonstrate no correlation between expected information gain and average PSNR. We also introduce an oracle baseline which directly observes the PSNR of each render and represents an ideal ordering. We compare FisherRF and D-Opt. (Block) with the baselines, where the best performing method is the method most similar to the oracle demonstrating a relationship between uncertainty and render quality. 

Fig. \ref{fig:blender_comparison_correlation} details the remaining plots for all objects in the dataset. In the main text, we chose figures which highlighted the performance of FisherRF as well as D-Opt. (Block). D-Opt. (Block) generally has a monotomic behavior and performance near the oracle. However, FisherRF sometimes does not exhibit the same behavior depending on the object. We note that this plot is not the intended goal of FisherRF, however we would expect a strong correlation between information gain and reconstruction error as stated previously.

\end{document}